\pdfoutput=1

\documentclass[table, x11names]{article} 
\usepackage{iclr2025_conference,times}

\usepackage{amsmath,amsfonts,bm}

\def\eqref#1{equation~\ref{#1}}

\def\1{\bm{1}}

\DeclareMathAlphabet{\mathsfit}{\encodingdefault}{\sfdefault}{m}{sl}
\SetMathAlphabet{\mathsfit}{bold}{\encodingdefault}{\sfdefault}{bx}{n}

\usepackage{hyperref}
\usepackage{url}
\usepackage{tcolorbox}

\usepackage{times}
\usepackage{latexsym}
\usepackage{amsmath}
\usepackage{mathtools}

\usepackage[T1]{fontenc}

\usepackage[utf8]{inputenc}

\usepackage{xcolor}
\usepackage{microtype}
\newtheorem{definition}{Definition}
\newtheorem{exmp}{Example}[section]
\usepackage{latexsym}
\usepackage{multicol, multirow}
\usepackage{booktabs}
\usepackage{arydshln}
\usepackage{hyperref}       
\usepackage{url}            
\usepackage{booktabs}       
\usepackage{amsfonts}       
\usepackage{nicefrac}       
\usepackage{microtype}      
\usepackage{lipsum}
\usepackage{graphicx}
\usepackage{enumitem}
\usepackage{color}
\usepackage{verbatimbox}
\usepackage{listings}

\usepackage{tikz}
\usetikzlibrary{intersections}
\usetikzlibrary{positioning}
\usepackage{wrapfig}

\definecolor{codegreen}{rgb}{0,0.6,0}
\definecolor{codegray}{rgb}{0.5,0.5,0.5}
\definecolor{codepurple}{rgb}{0.58,0,0.82}
\definecolor{backcolour}{rgb}{0.96,0.96,0.96}

\setlist[itemize]{leftmargin=*}
\setlist[enumerate]{leftmargin=*}
\graphicspath{ {./images/} }

\lstdefinestyle{mystyle}{
  backgroundcolor=\color{backcolour}, commentstyle=\color{codegreen},
  keywordstyle=\color{magenta},
  numberstyle=\tiny\color{codegray},
  stringstyle=\color{codepurple},
  basicstyle=\ttfamily\footnotesize,
  breakatwhitespace=false,         
  breaklines=true,                 
  captionpos=b,                    
  keepspaces=true,                 
  numbers=left,                    
  numbersep=5pt,                  
  showspaces=false,                
  showstringspaces=false,
  showtabs=false,                  
  tabsize=2
}

\usepackage[strict]{changepage}
\usepackage{framed}
\definecolor{demonstrationshade}{rgb}{0.95,0.95,1}
\definecolor{promptshade}{rgb}{0.95,0.95,1}

\definecolor{tea_green}{RGB}{214, 234, 193}
\definecolor{hint_green}{RGB}{226,246,209}
\definecolor{Madang}{RGB}{190,235,159}
\definecolor{yellow_green}{RGB}{198,222,119}
\definecolor{link_water}{RGB}{221, 232, 250}
\definecolor{celestial_blue}{RGB}{52, 152, 219}
\definecolor{shakespeare}{RGB}{85, 154, 193}
\definecolor{buttermilk}{RGB}{255,242,174}
\definecolor{chardonnay}{RGB}{250,196,114}
\definecolor{rajah}{RGB}{253,180,98}
\definecolor{fog}{RGB}{213, 193, 234}
\definecolor{melon}{RGB}{254,191,181}
\definecolor{sundown}{RGB}{249, 180, 181}
\definecolor{mona_lisa}{RGB}{246,152,134}
\definecolor{salmon}{RGB}{242,131,107}

\definecolor{saltpan}{RGB}{238, 243, 232}
\definecolor{aqua_spring}{RGB}{232, 243, 232}
\definecolor{tea_green}{RGB}{214, 234, 193}
\definecolor{Madang}{RGB}{190,235,159}
\definecolor{fringy_flower}{RGB}{194, 234, 193}
\definecolor{aero_blue}{RGB}{193, 234, 213}
\definecolor{pixie_green}{RGB}{183,214,170}
\definecolor{french_pass}{RGB}{195,232,246}
\definecolor{ice_cold}{RGB}{169,232,220}
\definecolor{pale_turquoise}{RGB}{172,240,242}
\definecolor{cruise}{RGB}{179,226,205}
\definecolor{sail}{RGB}{163,205,235}
\definecolor{spindle}{RGB}{179,205,227}
\definecolor{link_water}{RGB}{221, 232, 250}
\definecolor{periwinkle}{RGB}{203,213,232}
\definecolor{zanah}{RGB}{220, 233, 213}
\definecolor{frostee}{RGB}{217, 231, 214}
\definecolor{opal}{RGB}{199, 221, 211}
\definecolor{jet_stream}{RGB}{188, 214, 210}
\definecolor{skeptic}{RGB}{153, 187, 167}
\definecolor{hint_green}{RGB}{226,246,209}
\definecolor{snow_flurry}{RGB}{230,245,201}
\definecolor{surf_crest}{RGB}{205,230,208}
\definecolor{yellow_green}{RGB}{198,222,119}
\definecolor{cream}{RGB}{255,255,204}
\definecolor{pale_prim}{RGB}{255,255,179}
\definecolor{spring_sun}{RGB}{242,243,195}
\definecolor{portafino}{RGB}{245,237,160}
\definecolor{buttermilk}{RGB}{255,242,174}
\definecolor{cream_brulee}{RGB}{255, 229, 151}
\definecolor{dairy_cream}{RGB}{254,226,189}
\definecolor{champagne}{RGB}{254,217,166}
\definecolor{chardonnay}{RGB}{250,196,114}
\definecolor{manhattan}{RGB}{226,180,125}
\definecolor{rajah}{RGB}{253,180,98}
\definecolor{early_dawn}{RGB}{252,243,218}
\definecolor{egg_shell}{RGB}{238, 234, 215}
\definecolor{selago}{RGB}{243, 232, 243}
\definecolor{quartz}{RGB}{219,223,238}
\definecolor{fog}{RGB}{213, 193, 234}
\definecolor{languid_lavender}{RGB}{222,203,228}
\definecolor{watusi}{RGB}{254,221,207}
\definecolor{coral_andy}{RGB}{243,204,205}
\definecolor{cosmos}{RGB}{248,209,210}
\definecolor{melon}{RGB}{254,191,181}
\definecolor{azalea}{RGB}{234, 193, 194}
\definecolor{beauty_bush}{RGB}{235, 185, 179}
\definecolor{sundown}{RGB}{249, 180, 181}
\definecolor{mona_lisa}{RGB}{246,152,134}
\definecolor{salmon}{RGB}{242,131,107}

\definecolor{summer_sky}{RGB}{58, 151, 233}
\definecolor{chateau_green}{RGB}{72, 179, 96}
\definecolor{matisse}{RGB}{25, 104, 167}
\definecolor{allports}{RGB}{31, 106, 125}
\definecolor{sun_shade}{RGB}{255, 144, 68}
\definecolor{flamingo}{RGB}{237, 88, 85}
\definecolor{studio}{RGB}{128, 91, 160}

\definecolor{maya_blue}{RGB}{102, 204, 255}
\definecolor{feijoa}{RGB}{178,223,138}
\definecolor{sushi}{RGB}{117, 168, 47}
\definecolor{norway}{RGB}{158, 194, 132}
\definecolor{japanese_laurel}{RGB}{53, 116, 40}
\definecolor{see_green}{RGB}{161,228,195}
\definecolor{monte_carlo}{RGB}{135,204,194}
\definecolor{granny_smith_apple}{RGB}{150,214,150}
\definecolor{moss_green}{RGB}{170,216,176}
\definecolor{chateau_green}{RGB}{72, 179, 96}
\definecolor{opal}{RGB}{164,207,190}
\definecolor{acapulco}{RGB}{117, 170, 148}
\definecolor{viridian}{RGB}{55, 137, 122}
\definecolor{amazon}{RGB}{56, 123, 84}
\definecolor{asparagus}{RGB}{123, 160, 91}
\definecolor{fruit_salad}{RGB}{91, 160, 94}
\definecolor{puerto_rico}{RGB}{72, 179, 150}
\definecolor{mountain_meadow}{RGB}{0, 163, 136}
\definecolor{matisse}{RGB}{25, 104, 167}
\definecolor{allports}{RGB}{31, 106, 125}
\definecolor{astral}{RGB}{55, 111, 137}
\definecolor{spring_leaves}{RGB}{46, 83, 117}
\definecolor{biscay}{RGB}{44, 62, 80}
\definecolor{midnight}{RGB}{0, 29, 50}
\definecolor{amethyst}{RGB}{153, 102, 204}
\definecolor{studio}{RGB}{128, 91, 160}
\definecolor{tapestry}{RGB}{194, 109, 132}
\definecolor{atomic_tangerine}{RGB}{255, 153, 102}
\definecolor{amber}{RGB}{255, 191, 0}
\definecolor{casablanca}{RGB}{244, 178, 84}
\definecolor{california}{RGB}{233, 140, 58}
\definecolor{tomato}{RGB}{255, 97, 56} 
\definecolor{alizarin}{RGB}{233, 58, 64}

\definecolor{linen}{RGB}{251, 239, 227}
\definecolor{double_pearl_lusta}{RGB}{253, 242, 208}
\definecolor{oasis}{RGB}{253, 242, 208}
\definecolor{milan}{RGB}{255, 254, 169}
\definecolor{texas}{RGB}{245, 232, 123}
\definecolor{maize}{RGB}{249, 212, 156}

\definecolor{turmeric}{RGB}{211, 178, 76}
\definecolor{saffron}{RGB}{249,193,62}
\definecolor{my_sin}{RGB}{255, 176, 59}
\definecolor{tree_poppy}{RGB}{246, 154, 27}
\definecolor{jaffa}{RGB}{240, 131, 58}
\definecolor{crusta}{RGB}{254, 127, 44}
\definecolor{tahiti_gold}{RGB}{223, 102, 36}
\definecolor{outrageous_orange}{RGB}{255, 100, 45}
\definecolor{safety_orange}{RGB}{254, 106, 0}

\definecolor{azalea}{RGB}{251, 196, 196}
\definecolor{oyster_pink}{RGB}{238,206,205} 
\definecolor{coral_candy}{RGB}{242,208,205} 
\definecolor{baby_pink}{RGB}{246, 194, 192}
\definecolor{petite_orchid}{RGB}{223, 157, 155}
\definecolor{apricot}{RGB}{241,140,122}
\definecolor{NY_pink}{RGB}{228,136,113}
\definecolor{carmine_pink}{RGB}{231, 76, 60}
\definecolor{deep_carmine_pink}{RGB}{236, 50, 67}

\definecolor{wewak}{RGB}{244, 143, 150}
\definecolor{light_coral}{RGB}{244, 127, 123}
\definecolor{bittersweet}{RGB}{255,111,105}
\definecolor{carnation}{RGB}{245, 80, 86}
\definecolor{flamingo}{RGB}{237, 88, 85}
\definecolor{sunset_orange}{RGB}{242,89,75}
\definecolor{ku_crimson}{RGB}{243, 0, 25}
\definecolor{amaranth}{RGB}{234,46,73}
\definecolor{valencia}{RGB}{214, 87, 70}
\definecolor{chilean_fire}{RGB}{215, 87, 44}
\definecolor{mexican_red}{RGB}{170, 41, 37}

\definecolor{napa}{RGB}{163, 154, 137}

\definecolor{athens_gray}{RGB}{236, 240, 241}
\definecolor{gallery}{RGB}{240,240,240}
\definecolor{mercury}{RGB}{230,230,230}
\definecolor{platinum}{RGB}{228,228,228}
\definecolor{silver}{RGB}{191,191,191}
\definecolor{aluminum}{RGB}{153,153,153}
\definecolor{ship_gray}{RGB}{77,77,77}
\definecolor{tuatara}{RGB}{67, 67, 67}

\definecolor{malibu}{RGB}{110, 180, 240}
\definecolor{celestial_blue}{RGB}{52, 152, 219}
\definecolor{curious_blue}{RGB}{41, 128, 185}
\definecolor{french_blue}{RGB}{0, 112, 182}
\definecolor{matisse}{RGB}{25, 104, 167}
\definecolor{shakespeare}{RGB}{85, 154, 193}
\definecolor{seagull}{RGB}{128,177,211}
\definecolor{jelly_bean}{RGB}{45, 126, 150}
\definecolor{venice_blue}{RGB}{87, 135, 105}
\definecolor{boston_blue}{RGB}{68, 147, 161}

\definecolor{turquoise}{RGB}{41,217,194}
\definecolor{java}{RGB}{2,190,196}
\definecolor{riptide}{RGB}{141,211,199}
\definecolor{mountain_meadow}{RGB}{0, 163, 136}
\definecolor{free_speech_aquamarine}{RGB}{0, 156, 114}

\definecolor{cosmic_latte}{RGB}{222, 247, 229}
\definecolor{chinook}{RGB}{163, 232, 178}
\definecolor{padua}{RGB}{121, 189, 143}
\definecolor{ocean_green}{RGB}{79, 176, 112}
\definecolor{pastel_green}{RGB}{107, 227, 135}
\definecolor{chateau_green}{RGB}{69, 191, 85}
\definecolor{RoyalBlue}{RGB}{69, 191, 85}
\definecolor{pigment_green}{RGB}{0, 175, 79}
\definecolor{fern}{RGB}{101,197,117}
\definecolor{killarney}{RGB}{56, 113, 66}

\usepackage{pgfplots}
\usepgfplotslibrary{groupplots}
\usetikzlibrary{decorations.pathreplacing}
\pgfplotsset{
axis background/.style={fill=gallery},
grid=both,
  xtick pos=left,
  ytick pos=left,
  tick style={
    major grid style={style=white,line width=1pt},
    minor grid style=gallery,
    draw=none,
  },
  minor tick num=1,
}

\title{InductionBench: LLMs Fail in the Simplest\\Complexity Class}

\author{
  Wenyue Hua$^{1}$ Tyler Wong$^1$ Fei Sun\\
  Liangming Pan$^2$ Adam Jardine$^3$ William Yang Wang$^1$\footnote{Corresponding authors: wenyuehua@ucsb.edu, william@cs.ucsb.edu. I'm very grateful for extensive discussion with Wenda Xu, Xinyi Wang at UCSB.} \\
  \\
  $^1$University of California, Santa Barbara,\\
  $^2$University of Arizona,
  $^3$Rutgers University, New Brunswick
}

\begin{document}
\maketitle

\begin{abstract}
Large language models (LLMs) have shown remarkable improvements in reasoning and many existing benchmarks have been addressed by models such as o1 and o3 either fully or partially. However, a majority of these benchmarks emphasize deductive reasoning, including mathematical and coding tasks in which rules such as mathematical axioms or programming syntax are clearly defined, based on which LLMs can plan and apply these rules to arrive at a solution. In contrast, \textit{inductive reasoning}, where one infers the underlying rules from observed data, remains less explored. Such inductive processes lie at the heart of scientific discovery, as they enable researchers to extract general principles from empirical observations. To assess whether LLMs possess this capacity, we introduce \textbf{InductionBench}, a new benchmark designed to evaluate the inductive reasoning ability of LLMs. Our experimental findings reveal that even the most advanced models available struggle to master the simplest complexity classes within the subregular hierarchy of functions, highlighting a notable deficiency in current LLMs' inductive reasoning capabilities. Coda and data are available \url{https://github.com/Wenyueh/inductive_reasoning_benchmark}.
\end{abstract}

\section{Introduction}
The remarkable progress of large language models (LLMs) in recent years has yielded substantial improvements in their reasoning capabilities. This progress is most evident in benchmarks involving complex mathematics \citep{cobbe2021training, hendrycks2021measuring} and coding tasks \citep{jain2024livecodebench, jimenez2023swe, chen2021evaluating, fan2023nphardeval}. Beyond these domains, researchers have also explored the logical reasoning abilities of LLMs from various angles, including propositional logic \citep{zhu2023dyval}, first-order logic \citep{han2022folio, parmar2024logicbench}, and propositional logic under different contexts \citep{hua2024disentangling}.



\begin{wrapfigure}{r}{0.5\textwidth}
  \begin{center}
    \includegraphics[scale=0.35]{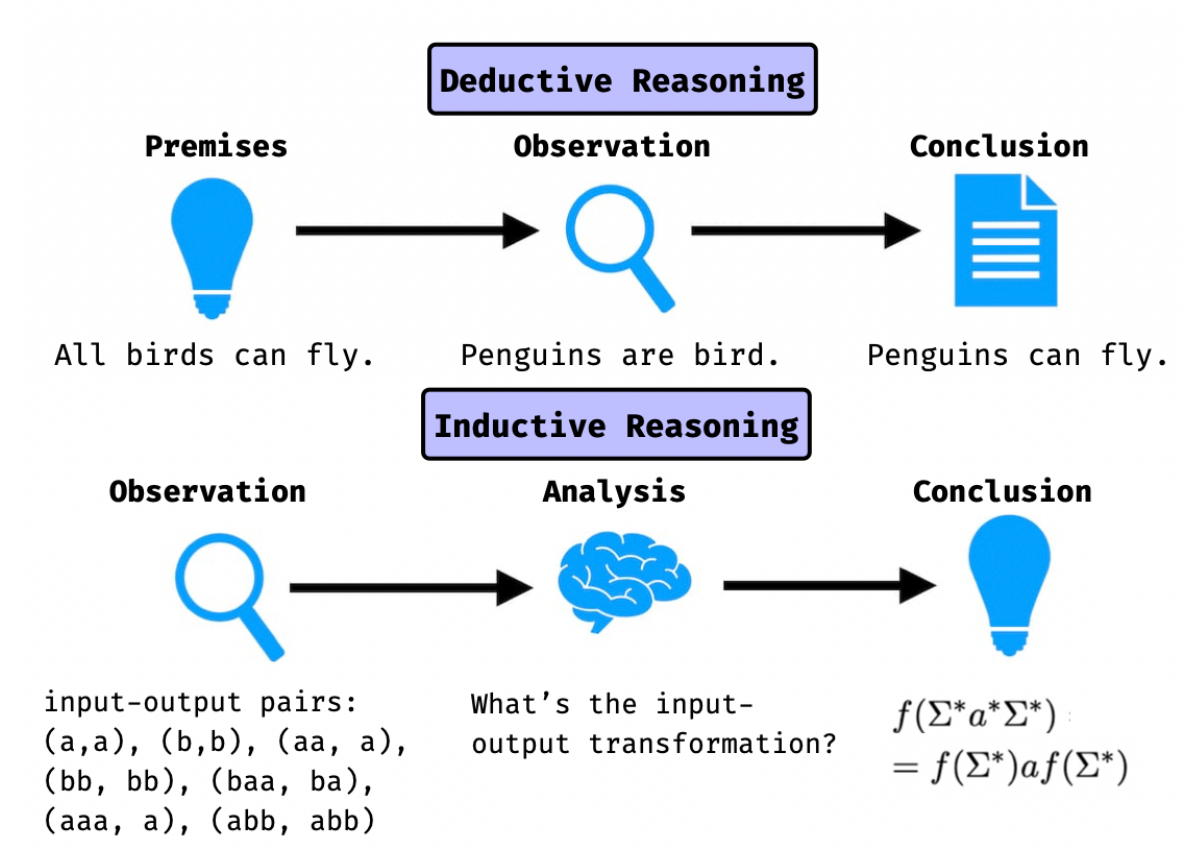}
    \vspace{-10pt}
    \caption{Deductive vs. Inductive Reasoning}
    \label{fig:demonstration}
  \end{center}
\end{wrapfigure}

Despite significant progress in model capabilities, existing benchmarks predominantly focus on deductive reasoning, largely overlooking inductive reasoning. The former requires applying explicitly defined premises to derive valid conclusions, whereas the latter requires inferring the underlying principles, rules, or patterns from observations \citep{hawthorne2004inductive}. Both forms of reasoning are essential; inductive reasoning, in particular, is critical in domains such as scientific discovery where researchers seek to characterize natural laws based on empirical data \citep{grunwald2007minimum, hansen2001model} that captures complex phenomena. Figure~\ref{fig:demonstration} illustrates the differences between inductive and deductive reasoning.

In this paper, we address this gap by introducing \textbf{InductionBench}, a rigorous benchmark designed to assess LLMs' inductive reasoning abilities by testing whether they can infer a string-to-string transformation from a finite set of input–output pairs. A model must hypothesize the underlying relationship between inputs and outputs based on a finite set of examples and then extrapolate those rules to unseen strings. The process of discovering the underlying function from limited data reflects the core principles of inductive reasoning.

\begin{wrapfigure}{r}{0.5\textwidth}
    \centering
\resizebox{0.9\linewidth}{!}{
    \begin{tikzpicture} 

  \draw[thick, draw=black] 
       (0,0) ellipse (6 and 3.6);
  \node at (0,3) {Regular Functions};

  \draw[
    thick, 
    draw=black, 
    fill=none, 
    fill opacity=0.3
  ] 
  (-2,0) ellipse (4 and 2.2);
  \node[] at (-2.4, 1.65) {Left Subsequential};

  \draw[
    thick, 
    draw=black, 
    fill=none, 
    fill opacity=0.3
  ] 
  (2,0) ellipse (4 and 2.2);
  \node[] at (2.4, 1.65) {Right Subsequential};

  \draw[
    thick, 
    fill=gray, 
    fill opacity=0.3
  ] 
  (-2,0) ellipse (2.8 and 1.2);
  \node[] at (-3.2, 0) {Left OSL};

  \draw[
    thick, 
    fill=gray,
    fill opacity=0.3
  ] 
  (2,0) ellipse (2.8 and 1.2);
  \node[] at (3.2, 0) {Right OSL};

  \draw[
    thick, 
    fill=gray, 
    fill opacity=0.3
  ] 
  (0,0) ellipse (0.8 and 1.87);
  \node[] at (0,1.35) {\textbf{ISL}};

\end{tikzpicture} }
    \caption{Subregular hierarchy in string-to-string maps}
    \vspace{-10pt}
    \label{fig:subregular}
\end{wrapfigure}
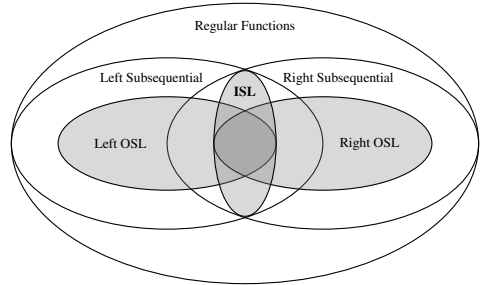

Our benchmark is grounded in the subregular hierarchy \citep{rogers2011aural, truthe2018hierarchy, graf2022diving, jager2012formal, heinz2018computational} (see Figure~\ref{fig:subregular}) for string-to-string mappings \citep{mohri1997finite}, focusing on input transformations restricted to regular functions. By systematically increasing task complexity across multiple classes in the subregular hierarchy, we gain detailed insights into how effectively LLMs detect, hypothesize, and generalize underlying rules from theoretically sufficient datapoints. 

We evaluate multiple state-of-the-art LLMs to understand LLM's inductive reasoning ability and identify factors that increase the difficulty of inductive reasoning tasks for LLMs, such as the length of the minimum-length description, the number of datapoints, and in-context examples. Through extensive experiments, we find that even advanced models such as o3-mini struggle with basic inductive tasks, highlighting a significant shortcoming in the current generation of LLMs. More detailed findings are presented in Section 5.

\section{Related Work}


\paragraph{Deductive Reasoning.} One major branch of reasoning benchmarks centers on deductive inference, where models apply established premises to derive specific conclusions. Notable examples include ReClor \citep{yu2020reclor}, which evaluates the ability to solve logical reasoning questions resembling those found in standardized tests, and various logic-based benchmarks of increasing complexity, from propositional logic to first-order logic \citep{han2022folio, parmar2024logicbench, zhu2023dyval, hua2024disentangling}. These tasks typically require handling structured logical relationships with minimal ambiguity in how premises lead to conclusions.

Another type of reasoning benchmarks is mathematical problem solving, including elementary arithmetic to advanced competition-level questions.  \citet{hendrycks2021measuring} test both computational skills and the sequential reasoning steps involved in mathematics. \citet{cobbe2021training} covers a broad spectrum of topics, including geometry and higher-level problem solving. However, most standard mathematics problem-solving tasks can be framed as deductive reasoning, as they involve applying established axioms, definitions, and theorems in a logically valid sequence to derive a conclusion. 

\paragraph{Inductive Reasoning.} Despite the diversity of existing benchmarks, inductive reasoning, where models hypothesize and generalize patterns from examples without pre-specified rules, remains comparatively underexplored. Current evaluations of inductive skills have largely been limited to small-scale symbolic regression, artificial language translation, and concept learning \citep{liu2024incomplete, lake2019human, qiu2023phenomenal}, which, although important in real-world scenarios, often lack three key elements: (1) an explicit analysis of the inherent difficulty of the task (2) a guarantee that the provided input–output dataset can identify the target function (3) a mechanism to evaluate whether models can identify the ``best possible hypothesis'' under Occam's Razor \citep{blumer1987occam, baker2007occam} principle, \emph{i.e.}, a description with minimal length \citep{hansen2001model, grunwald2007minimum}.

\paragraph{Our Contribution.} To address these shortcomings, we introduce a new benchmark targeting on inductive reasoning skills. Building on subregular hierarchy and corresponding polynomial time and data learnability guarantees, our benchmark, \textbf{InductionBench}, tests how effectively LLMs infer underlying transformation functions from finite datapoints. We also measure the degree to which models produce minimal, non-redundant hypotheses, providing a lens into their ability of generalization. Through a fine-grained, gradually increasing level of complexity, our evaluations reveal how current LLMs cope with the growing search space. There are several advantages of our benchmark:
\begin{enumerate}
    \item \textbf{Automated Evaluation}: Because the data is derived from well-defined functions, one can directly compare the model's output with the known ground-truth function, eliminating the need for expensive human annotations. 
    \item \textbf{Dynamic Data Generation}: The dataset is produced randomly based on specific function classes, allowing periodic ``refreshes'' to prevent models from relying on memorized examples.
    \item \textbf{Rigorous Assessment of Hypothesis Space}: As the function is well-defined, one can control the size of the hypothesis space with precision. This control enables a rigorous and systematic evaluation of LLM performance from a theoretically grounded perspective.
\end{enumerate}

\section{Computational Complexity in Inductive Reasoning}
\label{sec:com}
\textbf{InductionBench} uses string-to-string transformation/functions as a proxy to study inductive reasoning, which has established computational complexity hierarchy\citep{roche1997finite, engelfriet2001mso}. We focus on the subregular hierarchy, the hierarchy under regular functions. Though with limited expressive power, our experiments show that these classes already present substantial challenges for LLMs.

Specifically, we limit our attention to three classes of deterministic regular functions—\emph{Left Output-Strictly-Local} (L-OSL), \emph{Right Output-Strictly-Local} (R-OSL), and \emph{Input-Strictly-Local} (ISL), whose positions in the subregular hierarchy are illustrated in Figure~\ref{fig:subregular} \citep{heinz2018computational}. These classes represent
the lowest-complexity tier for string-to-string mappings within the subregular hierarchy. They are proper subclasses of subsequential function class and, more broadly, of weakly-deterministic class and non-deterministic class, which are themselves subsets of the regular function classes. Although we do not elaborate on the complete regular function hierarchy here, it is important to note that the ISL, L-OSL, and R-OSL classes are among the simplest in this framework.

Strictly local functions can be seen as operating with a fixed amount of look-ahead, similar to Markov processes. They are \emph{provably learnable in polynomial time from polynomially sized samples} \citep{chandlee2014learning, de1997characteristic, chandlee2015output, jardine2014very}. Moreover, prior work has shown that an algorithm exists to learn the unique (up to isomorphism) smallest subsequential finite-state transducer that represents such ISL, L-OSL, R-OSL functions \citep{satta1997string, arasu2009learning}. This property allows us to evaluate not only whether LLMs can discover the correct patterns but also whether they can identify the simplest or most concise representation consistent with the data. 

\subsection{Preliminary}
Before providing the definitions of the three function classes, we first introduce the fundamental mathematical notations and formal definitions underpinning our discussion of string-to-string transformations and their properties.

Let $\Sigma$ be a finite alphabet. We denote by $\Sigma^*$ the set of all finite strings over $\Sigma$, and by $\Sigma^{\leq n}$ the set of all strings over $\Sigma$ of length at most $n$. The empty string is denoted as $\lambda$. The set of prefixes of a string $w$ is denoted as \textsc{Pref}($w$), defined as $\{p\in \Sigma^*\mid \exists s\in \Sigma^* s.t. w = ps\},$ and the set of suffixes of $w$ denoted as \textsc{Suff}($w$), defined as $\{s\in \Sigma^*\mid \exists p\in \Sigma^* s.t. w = ps\}.$ The longest common prefix of a set of strings $S$ is denoted as \textsc{lcp}($S$), defined as 
\begin{equation}
    p\in\cap_{w\in S}\textsc{Pref}(w) \text{ such as } \forall p'\in\cap_{w\in S}\textsc{Pref}(w), |p'| < |p|.
\end{equation}
For any function $f:\Sigma^*\to\Gamma^*$ and $w\in\Sigma^*$, let the tails of $w$ with respect to $f$ be defined as
\begin{equation}
    \textsc{tails}_f(w) = \{(y, v)\mid f(wy) = uv \text{ and }
    u = \textsc{lcp}(f(w\Sigma^*))\}.
\end{equation}
Intuitively, \textsc{tails}$_f(w)$ collects all possible continuations $(y, v)$ by appending $y$ to $w$. It summarizes how $f$ might extend beyond the partial input $w$. The total number of distinct tails across all strings in $\Sigma^*$ provides a measure of how many different non-trivial local transformation $f$ encodes.

\subsection{Function Class Definition}
Based on the concepts outlined above, we define the three function classes.
\begin{definition}[ISL]
A function f is ISL if there is a $k$ such that for all $u_1, u_2\in\Sigma^*$, if $\textsc{Suff}^{k-1}(u_1) = \textsc{Suff}^{k-1}(u_2)$, then $\textsc{tails}_f(u_1) = \textsc{tails}_f(u_2)$.
\end{definition}

\begin{figure}[!ht]
    \centering
    \includegraphics[width=0.5\linewidth]{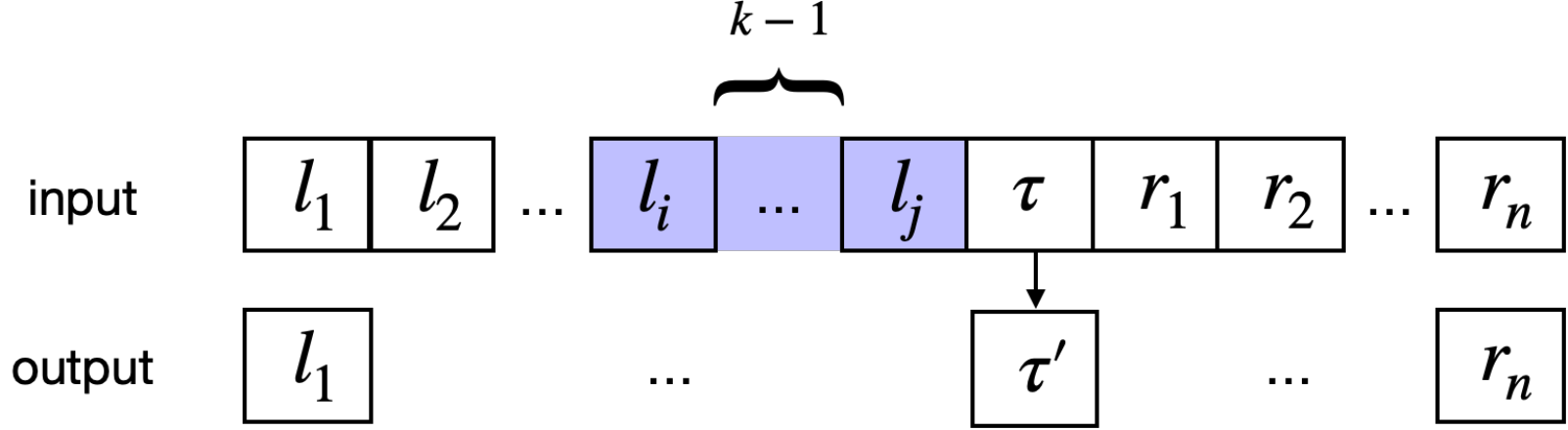}
    \caption{ISL definition}
    \label{fig:ISL}
\end{figure}

In simpler terms, this means that the output at each position in the string depends only on the preceding $k-1$ characters of the \emph{input}, making the transformation \emph{Markovian} with respect to the input. Figure\ref{fig:ISL} illustrates this definition. Below is a simple example:
\begin{exmp}
Suppose a function $f:\{a, b\}^*\to\{a, b\}^*$ rewrites each $b$ to $a$ \emph{only} if it appears after the input substring $ba$. In this scenario, we have $k=3$, and there are two distinct \emph{tails}:
\begin{equation*}
    \textsc{tails}_f(w) = \{(\lambda, \lambda),  (b, a), (bb, ab), (ab, ab) \dots\}, \quad 
    \forall w\in\Sigma^*\text{ such that } ba\in\textsc{suff}(w) 
\end{equation*}
and
\begin{equation*}
\textsc{tails}_f(w') = \{(\lambda, \lambda),  (a, a), (bb, bb), (ab, ab) \dots\}, \quad \forall w'\in\Sigma^*\text{ such that } ba\notin \textsc{suff}(w')
\end{equation*}
\end{exmp}

These tails indicate how the function's behavior shifts depending on whether the immediate context ends in $ba$. Such context-dependent tails also highlights that ISL functions can be effectively characterized or represented by local input constraints.

\begin{definition}[L-OSL]
A function f is L-OSL if there is a $k$ such that for all $u_1, u_2\in\Sigma^*$, if $\textsc{Suff}^{k-1}(f(u_1)) = \textsc{Suff}^{k-1}(f(u_2))$, then $\textsc{tails}_f(u_1) = \textsc{tails}_f(u_2)$.
\end{definition}

\begin{figure}[!ht]
    \centering
    \includegraphics[width=0.5\linewidth]{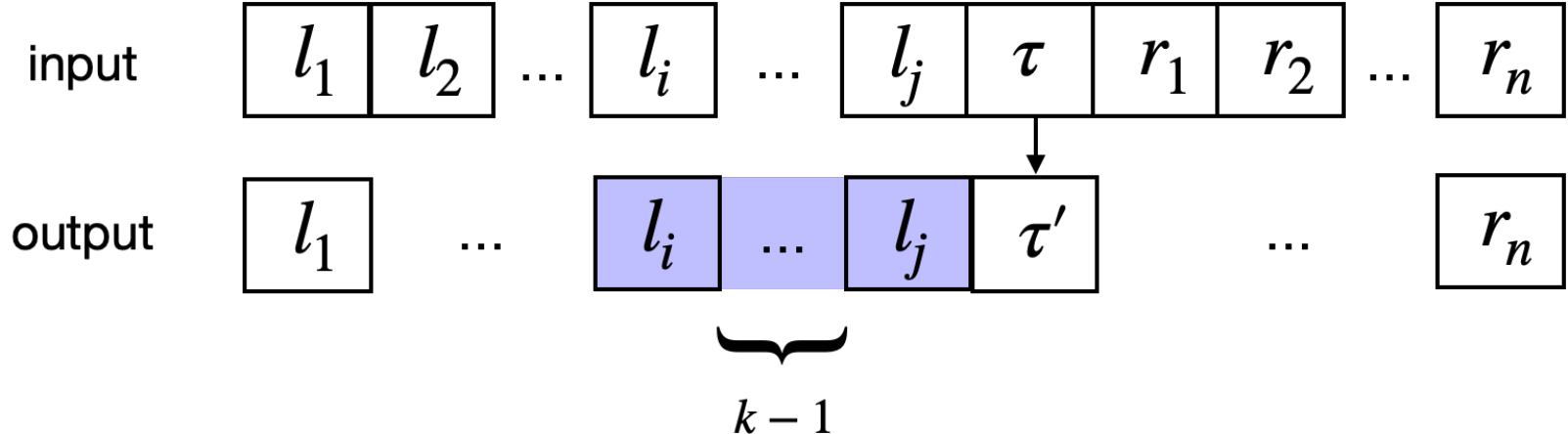}
    \caption{L-OSL definition}
    \label{fig:L_OSL}
\end{figure}

In other words, the output at each position in the transformed string depends only on the preceding $k-1$ characters of the \emph{output} itself, rather than on the input. This property can be understood as a form of Markovian process \emph{on the output}. Figure\ref{fig:L_OSL} illustrates this definition. Below is a simple example:
\begin{exmp}
Suppose a function $f$ rewrites each $b$ to $\lambda$ \emph{only} if it appears after the output substring $ba$. In this scenario, we have $k=3$, and there are two distinct \emph{tails}:
\begin{multline*}
    \textsc{tails}_f(w) = \{(\lambda, \lambda), (a, a), (b, \lambda),
    (bb, \lambda), (ab, ab), (ba, a), \dots\} \\
    \forall w\in\Sigma^*\text{ such that } ba\in\textsc{suff}(f(w))
\end{multline*}
and
\begin{multline*}
    \textsc{tails}_f(w) = \{(\lambda, \lambda), (a, a), (b, b),
    (bb, bb), (ab, ab), (ba, ba) \dots\} \\
    \forall w\in\Sigma^*\text{ such that } ba\notin\textsc{suff}(f(w))
\end{multline*}
\end{exmp}

\begin{figure}
    \centering
    \includegraphics[width=0.5\linewidth]{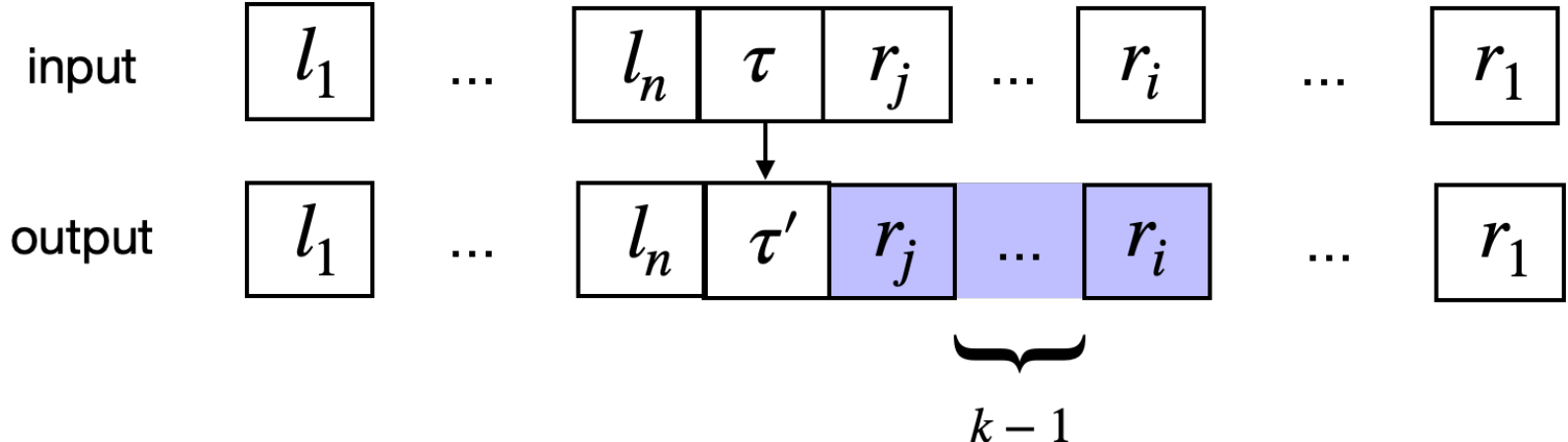}
    \caption{R-OSL definition}
    \label{fig:R_OSL}
\end{figure}

While L-OSL depends preceding output symbols to the ``left'', R-OSL functions depends on a limited number of \emph{future} output symbols to the ``right''. Conceptually, one can view R-OSL as analogous to L-OSL, except that the input is processed in reverse order. Although both belong to the broader OSL paradigm, they are \emph{incomparable} classes: each can express transformations the other cannot. The formal definition of R-OSL follows:

\begin{definition}[R-OSL]
A function f is R-OSL if there is a $k$ such that for all $u_1, u_2\in\Sigma^*$, if $\textsc{Suff}^{k-1}(f(u_1^{-1})) = \textsc{Suff}^{k-1}(f(u_2^{-1}))$, then $\textsc{tails}_f(u_1^{-1}) = \textsc{tails}_f(u_2^{-1})$.
\end{definition}

Intuitively, this class of functions can be viewed as a \emph{rightward} Markovian process on the output. Each output symbol is determined not by the preceding symbols as in L-OSL but by the next $k-1$ symbols that will appear in the output. Figure\ref{fig:L_OSL} illustrates this definition. 

The three classes, ISL, L-OSL, and R-OSL, are each deterministic and exhibit Markovian behavior, yet remain pairwise incomparable within the broader subregular hierarchy. In this work, we further restrict our attention to functions that involve \emph{substitution} which replaces one character with another and \emph{deletion} which maps a character to the empty string $\lambda$.

\subsection{Learnability}
The three function classes are \emph{identifiable in polynomial time using a polynomially sized characteristic sample} \citep{chandlee2014learning, chandlee2015output}. In other words, there exists a polynomial-time algorithm that, given sufficient data for a target function $f$, can produce a representation $\tau$ that satisfies $f(w) = \tau(w)$ for every $w\in\Sigma^*$. In other words, once sufficient data is presented, one can reliably recover a function equivalent to $f$ on all possible inputs. This learnability property underpins the value of these classes as testbeds for inductive reasoning, since the data requirement remains polynomial and successful inference is theoretically guaranteed.

We formalize ``sufficient data'' as the minimal set of input–output pairs needed to learn a $k$-strictly local function $f$, which is known as characteristic sample. Adapting the original definition\footnote{simplified from original definition} for clarity \citep{chandlee2014learning, chandlee2015output}, we define:
\begin{definition}[Characteristic Sample]
For a given $k$-ISL $f$, the characteristic sample $S$ is defined as $\{(w, w')\mid w\in \Sigma^{\leq k}\land f(w) = w'\}$. For a given $k$-OSL $f$, the characteristic sample $S$ is defined as $\{(w, w')\mid w'\in \Sigma^{\leq k}\land f(w) = w'\}$.
\end{definition}

If a provided dataset contains such characteristic sample, a learning algorithm can reconstruct a representation of $f$ that matches its behavior on every string in $\Sigma$. Accordingly, in the context of LLMs, we expect that providing this dataset as in-context examples should enable the model to induce the underlying string-to-string mapping. 

Below is an example characteristic sample with the transformation function being an ISL-3 function as described in Example 3.1: $f:\{a, b\}^*\to\{a, b\}^*$ rewrites each $b$ to $a$ \emph{only} if it appears after the input substring $ba$. 

\begin{minipage}[t]{1\linewidth} 
\begin{tcolorbox}[colback=gray!5,colframe=blue!40!black,title=Characteristic Sample for $f$]
\{($\lambda$, $\lambda$), ($a, a$), ($b, b$), ($aa, aa$), ($ab, ab$), ($ba, ba$), ($bb, bb$), ($aaa, aaa$), ($aab, aab$), ($aba, aba$), ($abb, abb$), ($baa, baa$), ($bab, ba\textcolor{red}{a}$), ($bba, bba$), ($bbb, bbb$)\}
\end{tcolorbox} 
\end{minipage} 

\subsection{Unique Function Representation}
Beyond verifying that a model can accurately discover a function from data, we also investigate how succinctly the model describes its inferred rules. This aspect is of both theoretical and practical interest: a minimal or \emph{most concise} representation not only offers interpretability advantages but can also reflect the model's capacity for truly generalizable, rather than merely enumerative, learning. 

One function can be represented or written in a non-unique way. For instance, consider an ISL function $f_1$ with $k=2$ over $\Sigma = \{a, b\}$ that maps the input character $a$ to $b$ when it comes after $b$, that rewrites each $a$ to $b$ only if the preceding character is $b$, while leaving other substrings unchanged. One concise description is:
\begin{equation}
\hspace{-3mm} f_1(w) {=}\,
  \begin{cases}
 f_1(w_1)ba^{-1}f_1(aw_2), \\
 \quad\quad \parbox[t]{0.6\linewidth}{
        if $w_1$ ends with $b$ and 
        $w {=}w_1$ $aw_2$ for some $w_1, w_2 {\in} \Sigma^*$
      }\\
 w, \,\,\,\,\, \text{otherwise}
  \end{cases}
\hspace{-10pt} 
\end{equation}

An alternative yet more verbose description of the same function might redundantly enumerate multiple cases:
\begin{equation} 
\hspace{-3mm} f_1'(w) {=}
\!\!\begin{cases} 
f_1'(w_1)ba^{-1}f_1'(aw_2), \\
 \quad\quad \parbox[t]{0.6\linewidth}{
 if $w_1$ ends with $ab$ and $w = w_1aw_2$ for some $w_1, w_2$ $\in\Sigma^*$ } \\
f_1'(w_1)ba^{-1}f_1'(aw_2) \\
\quad\quad \parbox[t]{0.6\linewidth}{ 
 if $w_1$ ends with $bb$ and $w = w_1aw_2$ for some $w_1, w_2$ $\in\Sigma^*$ }\\
w,\,\,\,\,\, \text{otherwise}
\end{cases} 
\hspace{-10pt}
\end{equation}

Although these two representations encode the same function, the second contains repetitive conditions and fails to emphasize that the output of $f_1$ depends solely on the single preceding character instead of the penultimate character. 

Because these functions admit a \emph{unique} minimal representation (up to isomorphism) \citep{chandlee2014learning, oncina1991inductive}, we can directly compare the function produced by an LLM to the ground-truth minimal form. In doing so, we evaluate whether the model not only \emph{discovers} the correct transformation but also \emph{simplifies} it to the most parsimonious description possible—an essential indicator of robust inductive reasoning.

\subsection{Rule-based Representation}
To streamline the generation and parsing of function representations, we employ a simplified notation wherein each transformation is written as ``condition $\circ$ target character $\to$ output of the target character'' \citep{bird1994one}.  In this notation, the \emph{condition} represents the minimal substring needed to trigger a transformation, while any input substring not matching this condition remains unchanged. For instance, in the earlier example, this approach permits a concise notation $b\circ a\to b$, indicating that the input $a$ is mapped to $b$ when it comes after $b$; otherwise, the input string remains unaltered. This concise, rule-based format simplifies both the model's output generation (by reducing complex functional descriptions) and our subsequent evaluation, as the applicable transformations can be easily parsible and verified.

To demonstrate the simplicity of rule-based representation: given an ISL function $f_2$ with $k=2$, the input $a$ becomes $b$ when it comes after $b$ and two consecutive $a$s will be reduced to one single $a$. The minimal function representation is as below:
\begin{equation}
\hspace{-3mm} f_2(w) {=}\,
  \begin{cases}
 f_2(w_1)ba^{-1}f_2(aw_2), \\
 \quad\quad \parbox[t]{0.6\linewidth}{
        if $w_1$ ends with $b$ and 
        $w {=}w_1aw_2$ for some $w_1, w_2 {\in} \Sigma^*$
      }\\
 f_2(w_1)a^{-1}f_2(aw_2), \\
 \quad\quad \parbox[t]{0.6\linewidth}{
        if $w_1$ ends with $a$ and 
        $w {=}w_1aw_2$ for some $w_1, w_2 {\in} \Sigma^*$
      }\\
 w, \,\,\,\,\, \text{otherwise}
  \end{cases}
\hspace{-10pt} 
\end{equation}

In the simplified rule-based format, it can be written as:$a\circ a\to\lambda\text{, }b\circ a\to b$. In summary, $f_1$ can be minimally expressed with a single rule, $f_2$ requires two rules.

\section{Benchmark Construction} 
In this section, we detail how our benchmark is constructed from the previously defined function classes. Each datapoint $(\mathcal{D}, f)$ in the benchmark is a pair of dataset $\mathcal{D}$ and function $f$ where $\mathcal{D}$ is a set of input-output pairs generated by $f$.

Each of ISL, L-OSL, and R-OSL classes can be further subdivided into incremental levels of complexity, determined by three key parameters: (1) the context window size $k$ (2) the vocabulary size $|\Sigma|$ (3) the minimal representation length of the function, \emph{i.e.} the minimal set of rules corresponding to the function. Given $k$ and $|\Sigma|$, the search space is $2^{|\Sigma|^k}$; given the number of rules $n$ additionally, the search space is $|\Sigma|^k \choose n$. To rigorously evaluate LLMs' inductive capabilities, we systematically vary these parameters across ISL, L-OSL, and R-OSL function classes.

In addition, we examine how performance changes with different numbers of input–output pairs in the prompt. Although having the characteristic sample present should theoretically guarantee recoverability of the underlying function, our empirical results indicate that the overall number of examples strongly affects performance. While extra data can provide richer information, it also increases context length considerably and heightens processing demands \citep{li2024long}. By varying the number of provided datapoints, we further investigate the extent to which the model engages in genuine reasoning and how robust its inductive abilities remain under changing input sizes.

\paragraph{Function Generation}
To systematically create benchmark instances, we first \emph{randomly generate} functions $f$ based on the three parameters: $k$, $|\Sigma|$, and the number of minimal rules describing $f$ by generating the set of rules that can describe $f$. While multiple representations of varying length can describe the same function, each function has a \emph{unique minimal representation} (up to isomorphism). During function generation, we therefore ensure that each function is expressed by a minimal, non-redundant rule set. Formally, if a $f$ is represented by a set of rules $R_f = \{r_1, r_2, ..., r_n\}$ where each $r_i$ has the form of $c_i\circ u_i\to v_i$ (with $c_i$ as the condition substring, $u_i$ the target character, and $v_i$ the transformed output for $u_i$), there are several constraints may be applied to functions belonging to the three classes. 

\begin{definition}[General Consistency]
Given $f$ represented by a set of rules $R_f: \forall r_i, r_j\in R_f, c_i\circ u_i\notin\textsc{Suff}(c_j\circ u_j)$ and $c_j\circ u_j\notin\textsc{Suff}(c_i\circ u_i)$.
\end{definition}

General Consistency ensures that the rules do not contradict one another or become redundant when conditions overlap. For instance, a function whose rule-based representation of $r_1: a\circ b\to a$ and $r_2: aa\circ b\to a$ is redundant, as the scenarios where $r_1$ is applied is a superset of the scenarios where $r_2$ is applied. For another instance, there does not exist a deterministic function that can be described by $r_1: a\circ b\to a$ and $r_2: aa\circ b\to \lambda$. Generating rule-based representations for ISL functions needs only satisfy this constraint.

\begin{definition}[OSL Non-Redundancy Guarantee]
Given $f$ represented by a set of rules $R:\forall r_i\in $ $R_f, \lnot\exists s_i'$ ${\in}\{s_i|s_i {\in} c_i\}$ such that $ \exists r_j\in R_f$ such that $ s_i' = c_j\circ u_j, \text{ unless }\exists r_k\in R$ such that $c_k\circ v_k = s_i'$.
\end{definition}

Constraint 2 is specific to the two OSL function classes because we need to make sure that all output conditions in the rule actually surface somewhere in the outputs of some datapoints. If the output condition $c$ never actually surface as the output, the rule will never be put into effect. Thereby the above rule basically requires that condition part of all rules can surface, either because it will never be modified by some other rule, or it emerges on the surface because of the application of other rule. For instance, a function represented by rules $r_1: aa\circ b\to a, r_2: a\circ a\to c$ is redundant because $r_1$ will never be applied because the string $aa$ will never surface as output and thus it will never be put into effect; For another instance, a function represented by $r_1: aa\circ b\to a, r_2: a\circ a\to c, r_3: a\circ d\to a$ is non redundant because even though into $aa$ string will be modified into $ac$, but $aa$ will surface in some datapoint because $ad$ will be modified into $aa$ and thus $r_1$ will be able to be applied.

Generating the functions following the two constraints, we ensure that the generated function representation is minimal, non-reducible guarantees a clear measure of complexity. One additional requirement is imposed to ensure each function indeed requires a look-ahead of size $k$. Specifically:

\begin{definition}[$k$-Complexity Guarantee]
Given $f$ whose designated context window $k = k_1$, $\exists r'\in R \text{ such that } c'\circ u'\to v'$ such that $|c'\circ u'| = k_1$.
\end{definition}

This condition guarantees that the function is genuinely $k$-strictly local (for ISL or OSL), rather than being representable with a smaller window size. Consequently, the functions we generate faithfully reflect the intended complexity level.

After generating the function $f$, we generate the characteristic sample of input-output pairs. For instance, given a function $f$ with $k = 2$ and $\Sigma = \{a, b\}$, the characteristic sample is $\{(a, f(a)),(b, f(b)), (ab, f(ab)), (aa, f(aa)), (bb, f(bb)), (ba, f(ba))\}$, a small set whose size is 6. By expanding this sample set, we can explore whether providing more than the minimal necessary examples aids or hinders the model's performance to infer the underlying function.

To evaluate how effectively an LLM can induce the underlying function, we include in the prompt (1) the function class, (2) context window $k$, (3) the alphabet $\Sigma$ which are information that guarantee learnability of the function. Then given the sample dataset, we request LLMs to produce a minimal rule-based description that reproduces the provided sample set, revealing whether it can \emph{discover} and \emph{optimally represent} the underlying transformation.

\section{Main Experiment}
\paragraph{Experiment Setting} We evaluate using zero-shot chain-of-thought prompting on six SOTA LLMs, including Llama-3.3-70b \citep{dubey2024llama} with FP8 quantization, Llama-3.1-405b with FP8 quantization, GPT-4o \citep{hurst2024gpt}, DeepSeek-V3 \citep{liu2024deepseek}, o1-mini\citep{jaech2024openai}, and o3-mini. For all models, we evaluate with all settings including $k\in\{2, 3, 4\}$, $|\Sigma|\in\{2, 3, 4\}$, number of rules $\in\{2, 3, 4\}$, and sample set size to be 1, 2, 3, 4 times larger than the characteristic sample. For each setting, we randomly generate 10 functions $f$ and corresponding input-output sample $\mathcal{D}$ to calculate the result. As o1-mini and o3-mini perform much better than other models, in addition, we evaluate on two more complex settings with $k\in\{4, 5\}, |\Sigma| = 5$.

\paragraph{Evaluation Metrics}
For each experiment setting, we leverage three metrics to evaluate performance: Precision, Recall, Compatibility. Let $R$ be the unique ground-truth rule set of minimal length for function $f$, $P$ be the predicted rule set generated by LLM, $\mathcal{D}$ be the provided sample set in the context on which we evaluate the correctness of $P$. \\
\textbf{Precision} measures how many of the predicted rules are correct relative to all rules the model generated: $\frac{|R\cap P|}{|P|}$. captures the proportion of the model's rules that align exactly with the ground-truth rules. A higher precision indicates fewer unnecessary/redundant rules.\\
\textbf{Recall} measures what fraction of the ground-truth rules the model successfully recovered: $\frac{|R\cap P|}{|R|}$. A higher recall reflects the model's ability to cover all aspects of the correct transformation.\\
\textbf{Compatibility} measures whether applying the predicted rule set $P$ to each input in the sample set $\mathcal{D}$ yields the correct output:
\begin{align*} 
\text{Compatibility}(P, \mathcal{D}) = 
\begin{cases} 
1 & \text{if } \forall (x_i, y_i)\in \mathcal{D}, P(x_i) = y_i\\
0 & \text{otherwise}
\end{cases} 
\end{align*}

Compatibility is the most fundamental measure, as it verifies whether the generated function accurately reproduces all observed input–output pairs in $\mathcal{D}$. A trivial way to achieve perfect compatibility is to include every pair $(x_i, y_i[-1])\in \mathcal{D}$ as an independent rule; however, doing so results in very low precision, indicating a failure to capture the underlying generalizable structure of the function. Note that all results presented are expressed as percentages.

\subsection{Model Input Prompt}
For each datapoint, the prompt provided to the LLM includes an informal definition of the function class that generated the data, indicating whether it belongs to ISL, L-OSL, or R-OSL, as well as the value of $k$. This information is included to simulate the prior knowledge typically assumed in algorithmic settings for learning such functions \cite{chandlee2014learning}. The sample of input-output pairs are attached after such prompt. We present the input prompt for datapoints under the function class of ISL below:

\begin{minipage}[t]{1\linewidth} 
\begin{tcolorbox}[colback=gray!5,colframe=blue!40!black,title=Input Prompt for ISL as Example]
Think step by step before providing the rules.\\
\\
Provide a minimum set of rules that describe the input-output transformations, \emph{i.e.} how an input character is transformed into its corresponding output character based on at most {{k-1}} input characters coming before it in the input.\\
Each rule describes how a character in the input is transformed into the output based on the context of the input characters preceding it.\\

For example: \\
a rule `abc --> a' means that the input character `c' is transformed into `a' in the output when it is preceded by `ab' in the input. 

Another example:\\
a rule `fqerrqb --> s' means that the input character `b' is transformed into `s' in the output when it is preceded by `fqerrq' in the input.\\

Therefore the left part of the rule basically describe the input context in which the input character is transformed into the output character.

Notice that the last character of the left part is the character to be transformed. 

The right part of the rule is the output character that the input character is transformed into. It can be a single character or a sequence of characters, it can also be empty string.\\

You don't need to provide trivial rules where input and output are the same, like `abdsa -> a' or `frqeb -> b'.
Notice that the left part of each rule always have length <= {{k}}. 

Write rules in the following format:\\
left -> right

Surround rules by XML tags <START> and <END> like below:

<START>\\
left1 -> right1\\
left2 -> right2\\
...\\
leftn -> rightn\\
<END>\\

THE RIGHT PART OF THE RULE SHOULD ONLY CONTAIN WHAT THE LAST CHARACTER OF THE LEFT PART IS TRANSFORMED INTO INSTEAD OF THE WHOLE STRING.

THINK STEP BY STEP BEFORE PROVIDING THE RULES.
\end{tcolorbox} 
\end{minipage}

\subsection{Model Performance Comparison}
\begin{table*}[!ht]
    \centering
    \renewcommand{\arraystretch}{1.1}
    \resizebox{14.5cm}{!}{
    \begin{tabular}{l ccc ccc ccc}
        \toprule
           \multirow{2.5}{*}{\bf Models} & \multicolumn{3}{c}{\bf ISL} & \multicolumn{3}{c}{\bf L-OSL} & \multicolumn{3}{c}{\bf R-OSL}\\
          \cmidrule(lr){2-4} \cmidrule(lr){5-7}  \cmidrule(lr){8-10}
           & recall  & precision & compatibility & recall  & precision & compatibility & recall  & precision & compatibility \\
         \midrule
           Llama-3.3 70B & 10.00 & 5.32 & 0.00 & 10.00 & 6.33 & 0.00 & 10.00 & 10.83 & 0.00\\
           Llama-3.1 405B & 10.00 & 3.75 & 0.00 & 6.67 & 1.10 & 0.00 & 13.33 & 1.85 & 0.00\\
           GPT-4o & 10.00 & 2.67 & 0.00 & 13.33 & 3.82 & 0.00 & 16.67 & 6.73 & 0.00\\
           DeepSeek-V3 & 13.33 & 2.46 & 0.00 & 23.33 & 2.73 & 0.00 & 3.33 & 0.25 & 0.00\\
           o1-mini & 36.67 & 22.09 & 0.00 & 43.33 & 32.12 & 0.00 & 26.67 & 17.58 & 0.00\\
           o3-mini & \textbf{73.33} & \textbf{59.58} & \textbf{10.00} & \textbf{66.67} & \textbf{69.17} & \textbf{10.00} & \textbf{63.33} & \textbf{62.00} & \textbf{30.00}\\
           \bottomrule
    \end{tabular}
    }
    \caption{Zero-shot CoT benchmark result with $k = 4$, $|\Sigma|$ = 4, number of rules = 3, sample size = 2}
    \vspace{-10pt}
    \label{tab:main}
\end{table*}
Table~\ref{tab:main} showcases model performance under particularly challenging conditions: $k=4, |\Sigma| = 4$ with 3 rules and sample size twice that of the minimal size of characteristic sample (detailed analysis on the impact of sample size is presented in Section 5.2). As seen in the table, compatibility scores collapse to 0 for all models except o3-mini, which also achieves relatively modest compatibility overall. This pattern highlights the difficulty that current LLMs face when required to track slightly broader contexts window even under very limited vocabulary size = 4. Full results are presented in Tables \ref{tab:ISL_main}, \ref{tab:LOSL_main}, \ref{tab:ROSL_main} in Appendix. Table~\ref{tab:harder} further reports the performance of o1-mini and o3-mini under slightly more challenging settings. Although both models generally exhibit non-trivial recall and precision, their compatibility scores consistently remain at or near zero. It is important to note that ISL, L-OSL, and R-OSL are the simplest function classes within the subregular hierarchy of string-to-string mappings. Thus, despite the strong performance of state-of-the-art models on benchmarks in coding \citep{jain2024livecodebench}, mathematics \citep{mirzadeh2024gsm}, and knowledge-intensive tasks \citep{wang2024mmlu}, they falter on this elementary inductive reasoning task. 

\begin{table}[!ht]
    \centering
    \begin{tabular}{l c ccc ccc}
        \toprule
          \multirow{2.5}{*}{\bf Models}& \multirow{2.5}{*}{\bf Settings}& \multicolumn{3}{c}{\bf k = 4} & \multicolumn{3}{c}{\bf k = 5}\\
          \cmidrule(lr){3-5} \cmidrule(lr){6-8} 
          & & R  & P & C & R  & P & C\\
            \midrule \multicolumn{8}{c}{\textbf{ISL}} \\
            \midrule
           \multirow{2}{*}{o1-mini} & rules = 4 & 25.00 & 12.21 & 0.00 & 10.00 & 9.10 & 0.00\\
           & rules = 5 & 36.00 & 40.41 & 0.00 & 10.00 & 3.14 & 0.00\\
           \hdashline
           \multirow{2}{*}{o3-mini} & rules = 4 & 37.50 & 49.83 & 0.00 & 27.50 & 30.75 & 0.00\\
           & rules = 5 & 42.00 & 58.67 & 0.00 & 20.00 & 38.33 & 0.00\\
           \midrule \multicolumn{8}{c}{\textbf{L-OSL}} \\
            \midrule
           \multirow{2}{*}{o1-mini} & rules = 4 & 32.50 & 37.34 & 0.00 & 15.00 & 29.33 & 10.00\\
           & rules = 5 & 28.00 & 23.17 & 0.00 & 8.00 & 4.61 & 0.00\\
           \hdashline
           \multirow{2}{*}{o3-mini} & rules = 4 & 57.50 & 58.93 & 0.00 & 22.50 & 39.26 & 0.00\\
           & rules = 5 & 48.00 & 71.38 & 0.00 & 10.00 & 23.67 & 0.00\\
           \midrule \multicolumn{8}{c}{\textbf{R-OSL}} \\
            \midrule
           \multirow{2}{*}{o1-mini} & rules = 4 & 17.50 & 22.33 & 0.00 & 12.50 & 7.63 & 0.00\\
           & rules = 5 & 16.00 & 15.42 & 0.00 & 18.00 & 22.82 & 0.00\\
           \hdashline
           \multirow{2}{*}{o3-mini} & rules = 4 & 45.00 & 43.76 & 10.00 & 20.00 & 50.00 & 0.00\\
           & rules = 5 & 38.00 & 55.17 & 0.00 & 14.00 & 36.17 & 0.00\\
           \bottomrule
    \end{tabular}
    \caption{o1-mini and o3-mini results (R = Recall, P = Precision, C = Compatibility) on harder setting with $k\in\{4, 5\}, |\Sigma| = 5$, sample size = 2}
    \vspace{-10pt}
    \label{tab:harder}
\end{table}

\subsection{Impact of Different Factors}
\begin{figure*}[!ht]
\centering

\resizebox{\textwidth}{!}{
    \begin{tikzpicture}
	\begin{groupplot}[
	    group style={group size=3 by 1,
	        horizontal sep = 64pt,
	        }, 
	    width=1\linewidth,
	     height=0.618\linewidth,
	     enlarge x limits=0.06,
	     ylabel=\huge recall,
	     ymin=0, ymax=100,
         xticklabels={rules=1, rules=2, rules=3, rules=1, rules=2, rules=3, rules=1, rules=2, rules=3},
	     xtick={1,2,3,4,5,6,7,8,9},
	     xticklabel style={font=\Large, rotate=-30},
	     yticklabel style={font=\Large},
	     ybar=1pt,
		every axis plot/.style={bar width=5pt},
	     ymajorgrids,
	     major grid style={draw=white},
	     y axis line style={opacity=0},
	     tickwidth=0pt,
	    ]
	    
		\nextgroupplot[
		legend style = {
		  font=\huge,
          draw=none, 
          draw opacity=0,
          fill=none,
          column sep = 2pt, 
          /tikz/every even column/.append style={column sep=5mm},
          legend columns = -1, 
          legend to name = grouplegend},
		title={\huge Vocab size = 2}, 
		after end axis/.append code={ 
		    \draw[decorate, decoration={brace, amplitude=10pt, mirror, raise=5pt}, 
		          very thick, black] 
		        (axis cs:0.6, -12) -- (axis cs: 3.4, -12)
		        node[midway, below=15pt, font=\huge] {k=2};
		    \draw[decorate, decoration={brace, amplitude=10pt, mirror, raise=5pt}, 
		          very thick, black] 
		        (axis cs:3.6, -12) -- (axis cs: 6.4, -12)
		        node[midway, below=15pt, font=\huge] {k=3};
		    \draw[decorate, decoration={brace, amplitude=10pt, mirror, raise=5pt}, 
		          very thick, black] 
		        (axis cs:6.6, -12) -- (axis cs: 9.4, -12)
		        node[midway, below=15pt, font=\huge] {k=4};
		}, 
		]
		\addplot [draw=none, fill=mona_lisa]
	     coordinates {
	       (1, 60) (2, 60) (3, 53.33) (4, 30) (5, 45) (6, 30) (7, 10) (8, 15) (9, 16.67)}; \addlegendentry{Llama-3.3 70B}
	     \addplot [draw=none, fill=flamingo]
	     coordinates {
	        (1, 30) (2, 55) (3, 56.67) (4, 50) (5, 10) (6, 10) (7, 20) (8, 10) (9, 0)}; \addlegendentry{Llama-3.1 405B}
	     \addplot [draw=none, fill=shakespeare]
	     coordinates {
	       (1, 90) (2, 80) (3, 70) (4, 50) (5, 40) (6, 40) (7, 50) (8, 15) (9, 33.33)}; \addlegendentry{DeepSeek-V3}
	     \addplot[draw=none, fill=sunset_orange] coordinates {
	      (1, 60) (2, 60) (3, 73.33) (4, 50) (5, 35) (6, 36.67) (7, 10) (8, 15) (9, 13.33)
	     };  \addlegendentry{GPT-4o}
	     \addplot[draw=none, fill=sushi] coordinates {
	        (1, 50) (2, 75) (3, 66.67) (4, 70) (5, 70) (6, 43.33) (7, 30) (8, 50) (9, 40)
	     }; \addlegendentry{o1-mini}

        \nextgroupplot[
        ylabel=\huge recall,
        after end axis/.append code={ 
		    \draw[decorate, decoration={brace, amplitude=10pt, mirror, raise=5pt}, 
		          very thick, black] 
		        (axis cs:0.6, -12) -- (axis cs: 3.4, -12)
		        node[midway, below=15pt, font=\huge] {k=2};
		    \draw[decorate, decoration={brace, amplitude=10pt, mirror, raise=5pt}, 
		          very thick, black] 
		        (axis cs:3.6, -12) -- (axis cs: 6.4, -12)
		        node[midway, below=15pt, font=\huge] {k=3};
		    \draw[decorate, decoration={brace, amplitude=10pt, mirror, raise=5pt}, 
		          very thick, black] 
		        (axis cs:6.6, -12) -- (axis cs: 9.4, -12)
		        node[midway, below=15pt, font=\huge] {k=4};
		}, 
        title={\huge Vocab size = 3}]
		 \addplot [draw=none, fill=mona_lisa]
	     coordinates {
	        (1, 70) (2, 85) (3, 66.67) (4, 20) (5, 10) (6, 33.33) (7, 20) (8, 5) (9, 6.67)}; 
	     \addplot [draw=none, fill=flamingo]
	     coordinates {
	         (1, 20) (2, 45) (3, 50) (4, 20) (5, 10) (6, 6.67) (7, 10) (8, 0) (9, 10)};
	     \addplot [draw=none, fill=shakespeare]
	     coordinates {
	        (1, 70) (2, 80) (3, 80) (4, 70) (5, 40) (6, 56.67) (7, 50) (8, 25) (9, 40)};
	     \addplot[draw=none, fill=sunset_orange] coordinates {
	      (2, 60) (3, 66.67) (4, 50) (5, 25) (6, 30) (7, 20) (8, 30) (9, 10)
	     };  
	     \addplot[draw=none, fill=sushi] coordinates {
	        (1, 80) (2, 90) (3, 80) (4, 50) (5, 40) (6, 63.33) (7, 30) (8, 55) (9, 30)
	     }; 

                \nextgroupplot[
        ylabel=\huge recall,
        after end axis/.append code={ 
		    \draw[decorate, decoration={brace, amplitude=10pt, mirror, raise=5pt}, 
		          very thick, black] 
		        (axis cs:0.6, -12) -- (axis cs: 3.4, -12)
		        node[midway, below=15pt, font=\huge] {k=2};
		    \draw[decorate, decoration={brace, amplitude=10pt, mirror, raise=5pt}, 
		          very thick, black] 
		        (axis cs:3.6, -12) -- (axis cs: 6.4, -12)
		        node[midway, below=15pt, font=\huge] {k=3};
		    \draw[decorate, decoration={brace, amplitude=10pt, mirror, raise=5pt}, 
		          very thick, black] 
		        (axis cs:6.6, -12) -- (axis cs: 9.4, -12)
		        node[midway, below=15pt, font=\huge] {k=4};
		}, 
        title={\huge Vocab size = 4}]
		 \addplot [draw=none, fill=mona_lisa]
	     coordinates {
	        (1, 60) (2, 40) (3, 53.33) (4, 30) (5, 15) (6, 6.67) (7, 10) (8, 0) (9, 10)}; 
	     \addplot [draw=none, fill=flamingo]
	     coordinates {
	          (1, 40) (2, 75) (3, 40) (4, 10) (5, 10) (6, 13.33) (7, 10) (8, 0) (9, 10)};
	     \addplot [draw=none, fill=shakespeare]
	     coordinates {
	         (1, 80) (2, 85) (3, 76.67) (4, 50) (5, 65) (6, 50) (7, 40) (8, 45) (9, 13.33)};
	     \addplot[draw=none, fill=sunset_orange] coordinates {
	       (1, 50) (2, 45) (3, 56.67) (4, 50) (5, 45) (6, 33.33) (7, 50) (8, 0) (9, 10)
	     };  
	     \addplot[draw=none, fill=sushi] coordinates {
	        (1, 80) (2, 75) (3, 93.33) (4, 60) (5, 50) (6, 46.67) (7, 40) (8, 35) (9, 36.67)
	     }; 

	\end{groupplot}
	


\node at ($(group c1r1) + (420pt, 160pt)$) {\pgfplotslegendfromname{grouplegend}};
\end{tikzpicture} }

\resizebox{\textwidth}{!}{
    \begin{tikzpicture}
	\begin{groupplot}[
	    group style={group size=3 by 1,
	        horizontal sep = 64pt,
	        }, 
	    width=1\linewidth,
	     height=0.618\linewidth,
	     enlarge x limits=0.06,
	     ylabel=\huge precision,
	     ymin=0, ymax=100,
         xticklabels={rules=1, rules=2, rules=3, rules=1, rules=2, rules=3, rules=1, rules=2, rules=3},
	     xtick={1,2,3,4,5,6,7,8,9},
	     xticklabel style={font=\Large, rotate=-30},
	     yticklabel style={font=\Large},
	     ybar=1pt,
		every axis plot/.style={bar width=5pt},
	     ymajorgrids,
	     major grid style={draw=white},
	     y axis line style={opacity=0},
	     tickwidth=0pt,
	    ]
	    
		\nextgroupplot[
		title={\huge Vocab size = 2}, 
		after end axis/.append code={ 
		    \draw[decorate, decoration={brace, amplitude=10pt, mirror, raise=5pt}, 
		          very thick, black] 
		        (axis cs:0.6, -12) -- (axis cs: 3.4, -12)
		        node[midway, below=15pt, font=\huge] {k=2};
		    \draw[decorate, decoration={brace, amplitude=10pt, mirror, raise=5pt}, 
		          very thick, black] 
		        (axis cs:3.6, -12) -- (axis cs: 6.4, -12)
		        node[midway, below=15pt, font=\huge] {k=3};
		    \draw[decorate, decoration={brace, amplitude=10pt, mirror, raise=5pt}, 
		          very thick, black] 
		        (axis cs:6.6, -12) -- (axis cs: 9.4, -12)
		        node[midway, below=15pt, font=\huge] {k=4};
		}, 
		]
		\addplot [draw=none, fill=mona_lisa]
coordinates {
    (1, 55.00) (2, 65.00) (3, 68.33) (4, 23.33) (5, 60.00) (6, 46.67) (7, 10.00) (8, 8.25) (9, 8.54)
}; 
\addplot [draw=none, fill=flamingo]
coordinates {
    (1, 25.00) (2, 50.00) (3, 44.17) (4, 35.00) (5, 6.67) (6, 9.50) (7, 15.00) (8, 7.50) (9, 0.00)
}; 
\addplot [draw=none, fill=shakespeare]
coordinates {
    (1, 60.00) (2, 60.00) (3, 54.83) (4, 32.50) (5, 19.89) (6, 26.15) (7, 12.33) (8, 3.82) (9, 10.61)
}; 
\addplot[draw=none, fill=sunset_orange] coordinates {
    (1, 43.33) (2, 37.50) (3, 68.33) (4, 22.00) (5, 18.43) (6, 19.30) (7, 2.00) (8, 5.63) (9, 2.50)
}; 
\addplot[draw=none, fill=sushi] coordinates {
    (1, 45.00) (2, 75.00) (3, 61.67) (4, 45.83) (5, 60.00) (6, 34.00) (7, 16.67) (8, 28.67) (9, 38.52)
}; 

        \nextgroupplot[
        ylabel=\huge precision,
        after end axis/.append code={ 
		    \draw[decorate, decoration={brace, amplitude=10pt, mirror, raise=5pt}, 
		          very thick, black] 
		        (axis cs:0.6, -12) -- (axis cs: 3.4, -12)
		        node[midway, below=15pt, font=\huge] {k=2};
		    \draw[decorate, decoration={brace, amplitude=10pt, mirror, raise=5pt}, 
		          very thick, black] 
		        (axis cs:3.6, -12) -- (axis cs: 6.4, -12)
		        node[midway, below=15pt, font=\huge] {k=3};
		    \draw[decorate, decoration={brace, amplitude=10pt, mirror, raise=5pt}, 
		          very thick, black] 
		        (axis cs:6.6, -12) -- (axis cs: 9.4, -12)
		        node[midway, below=15pt, font=\huge] {k=4};
		}, 
		title={\huge Vocab size = 3}
		]
        \addplot [draw=none, fill=mona_lisa]
		coordinates {
		    (1, 60.00) (2, 83.33) (3, 74.17) (4, 20.00) (5, 7.50) (6, 35.36) (7, 8.33) (8, 2.50) (9, 3.43)
		}; 
		\addplot [draw=none, fill=flamingo]
		coordinates {
		    (1, 10.00) (2, 32.58) (3, 38.45) (4, 10.00) (5, 7.50) (6, 3.10) (7, 10.00) (8, 0.00) (9, 5.27)
		}; 
		\addplot [draw=none, fill=shakespeare]
		coordinates {
		    (1, 65.00) (2, 56.00) (3, 60.76) (4, 45.00) (5, 13.23) (6, 21.64) (7, 13.93) (8, 1.89) (9, 5.88)
		}; 
		\addplot[draw=none, fill=sunset_orange] coordinates {
		    (1, 33.33) (2, 40.42) (3, 64.33) (4, 18.33) (5, 6.00) (6, 9.61) (7, 4.17) (8, 6.39) (9, 1.72)
		}; 
		\addplot[draw=none, fill=sushi] coordinates {
		    (1, 80.00) (2, 90.00) (3, 77.33) (4, 43.33) (5, 25.11) (6, 36.25) (7, 0.00) (8, 24.82) (9, 20.44)
		}; 

         \nextgroupplot[
        ylabel=\huge precision,
        after end axis/.append code={ 
		    \draw[decorate, decoration={brace, amplitude=10pt, mirror, raise=5pt}, 
		          very thick, black] 
		        (axis cs:0.6, -12) -- (axis cs: 3.4, -12)
		        node[midway, below=15pt, font=\huge] {k=2};
		    \draw[decorate, decoration={brace, amplitude=10pt, mirror, raise=5pt}, 
		          very thick, black] 
		        (axis cs:3.6, -12) -- (axis cs: 6.4, -12)
		        node[midway, below=15pt, font=\huge] {k=3};
		    \draw[decorate, decoration={brace, amplitude=10pt, mirror, raise=5pt}, 
		          very thick, black] 
		        (axis cs:6.6, -12) -- (axis cs: 9.4, -12)
		        node[midway, below=15pt, font=\huge] {k=4};
		}, 
        title={\huge Vocab size = 4}
        ]
		\addplot [draw=none, fill=mona_lisa]
		coordinates {
		    (1, 60.00) (2, 40.00) (3, 68.33) (4, 30.00) (5, 11.67) (6, 5.00) (7, 10.00) (8, 0.00) (9, 5.32)
		}; 
		\addplot [draw=none, fill=flamingo]
		coordinates {
		    (1, 35.00) (2, 52.33) (3, 34.33) (4, 5.00) (5, 5.83) (6, 1.54) (7, 10.00) (8, 0.00) (9, 3.75)
		}; 
		\addplot [draw=none, fill=shakespeare]
		coordinates {
		    (1, 52.50) (2, 57.15) (3, 63.12) (4, 17.00) (5, 18.66) (6, 16.05) (7, 14.58) (8, 5.30) (9, 2.46)
		}; 
		\addplot[draw=none, fill=sunset_orange] coordinates {
		    (1, 40.00) (2, 29.00) (3, 38.62) (4, 16.67) (5, 12.62) (6, 20.60) (7, 21.67) (8, 0.00) (9, 2.67)
		}; 
		\addplot[draw=none, fill=sushi] coordinates {
		    (1, 70.00) (2, 63.33) (3, 91.67) (4, 50.00) (5, 29.93) (6, 37.72) (7, 15.00) (8, 35.00) (9, 22.09)
		}; 

	\end{groupplot}

\end{tikzpicture} }

\resizebox{\textwidth}{!}{
    \begin{tikzpicture}
	\begin{groupplot}[
	    group style={group size=3 by 1,
	        horizontal sep = 64pt,
	        }, 
	    width=1\linewidth,
	     height=0.618\linewidth,
	     enlarge x limits=0.06,
	     ylabel=\huge compatibility,
	     ymin=0, ymax=100,
         xticklabels={rules=1, rules=2, rules=3, rules=1, rules=2, rules=3, rules=1, rules=2, rules=3},
	     xtick={1,2,3,4,5,6,7,8,9},
	     xticklabel style={font=\Large, rotate=-30},
	     yticklabel style={font=\Large},
	     ybar=1pt,
		every axis plot/.style={bar width=5pt},
	     ymajorgrids,
	     major grid style={draw=white},
	     y axis line style={opacity=0},
	     tickwidth=0pt,
	    ]
	    
		\nextgroupplot[
		title={\huge Vocab size = 2}, 
		after end axis/.append code={ 
		    \draw[decorate, decoration={brace, amplitude=10pt, mirror, raise=5pt}, 
		          very thick, black] 
		        (axis cs:0.6, -12) -- (axis cs: 3.4, -12)
		        node[midway, below=15pt, font=\huge] {k=2};
		    \draw[decorate, decoration={brace, amplitude=10pt, mirror, raise=5pt}, 
		          very thick, black] 
		        (axis cs:3.6, -12) -- (axis cs: 6.4, -12)
		        node[midway, below=15pt, font=\huge] {k=3};
		    \draw[decorate, decoration={brace, amplitude=10pt, mirror, raise=5pt}, 
		          very thick, black] 
		        (axis cs:6.6, -12) -- (axis cs: 9.4, -12)
		        node[midway, below=15pt, font=\huge] {k=4};
		}, 
		]
		\addplot [draw=none, fill=mona_lisa]
coordinates {
    (1, 60.00) (2, 50.00) (3, 20.33) (4, 20) (5, 30.00) (6, 10) (7, 10) (8, 0) (9, 0)
}; 
\addplot [draw=none, fill=flamingo]
coordinates {
    (1, 30.00) (2, 40.00) (3, 20) (4, 30) (5, 0) (6, 0) (7, 10.00) (8, 0) (9, 0)
}; 
\addplot [draw=none, fill=shakespeare]
coordinates {
    (1, 60.00) (2, 40.00) (3, 40) (4, 50) (5, 10.00) (6, 0) (7, 30) (8, 0) (9, 0)
}; 
\addplot[draw=none, fill=sunset_orange] coordinates {
    (1, 40.00) (2, 50) (3, 60) (4, 40) (5, 20) (6, 0) (7, 0) (8, 10) (9, 0)
}; 
\addplot[draw=none, fill=sushi] coordinates {
    (1, 50) (2, 75.00) (3, 60) (4, 40) (5, 40) (6, 0.00) (7, 10.00) (8, 0) (9, 0)
}; 

        \nextgroupplot[
        ylabel=\huge compatibility,
        after end axis/.append code={ 
		    \draw[decorate, decoration={brace, amplitude=10pt, mirror, raise=5pt}, 
		          very thick, black] 
		        (axis cs:0.6, -12) -- (axis cs: 3.4, -12)
		        node[midway, below=15pt, font=\huge] {k=2};
		    \draw[decorate, decoration={brace, amplitude=10pt, mirror, raise=5pt}, 
		          very thick, black] 
		        (axis cs:3.6, -12) -- (axis cs: 6.4, -12)
		        node[midway, below=15pt, font=\huge] {k=3};
		    \draw[decorate, decoration={brace, amplitude=10pt, mirror, raise=5pt}, 
		          very thick, black] 
		        (axis cs:6.6, -12) -- (axis cs: 9.4, -12)
		        node[midway, below=15pt, font=\huge] {k=4};
		}, 
		title={\huge Vocab size = 3}
		]
        \addplot [draw=none, fill=mona_lisa]
		coordinates {
		    (1, 60.00) (2, 60.00) (3, 20) (4, 20.00) (5, 0) (6, 0) (7, 10) (8, 0) (9, 0)
		}; 
		\addplot [draw=none, fill=flamingo]
		coordinates {
		    (1, 10) (2, 20) (3, 20) (4, 10) (5, 0) (6, 0) (7, 10.00) (8, 0.00) (9, 0)
		}; 
		\addplot [draw=none, fill=shakespeare]
		coordinates {
		    (1, 60) (2, 40.00) (3, 50) (4, 60) (5, 0) (6, 0) (7, 50) (8, 0) (9, 0)
		}; 
		\addplot[draw=none, fill=sunset_orange] coordinates {
		    (1, 50) (2, 30) (3, 40) (4, 40) (5, 0.00) (6, 0) (7, 10) (8, 0) (9, 0)
		}; 
		\addplot[draw=none, fill=sushi] coordinates {
		    (1, 80.00) (2, 80.00) (3, 60) (4, 40) (5, 10) (6, 10) (7, 0.00) (8,10) (9, 0)
		}; 

         \nextgroupplot[
        ylabel=\huge compatibility,
        after end axis/.append code={ 
		    \draw[decorate, decoration={brace, amplitude=10pt, mirror, raise=5pt}, 
		          very thick, black] 
		        (axis cs:0.6, -12) -- (axis cs: 3.4, -12)
		        node[midway, below=15pt, font=\huge] {k=2};
		    \draw[decorate, decoration={brace, amplitude=10pt, mirror, raise=5pt}, 
		          very thick, black] 
		        (axis cs:3.6, -12) -- (axis cs: 6.4, -12)
		        node[midway, below=15pt, font=\huge] {k=3};
		    \draw[decorate, decoration={brace, amplitude=10pt, mirror, raise=5pt}, 
		          very thick, black] 
		        (axis cs:6.6, -12) -- (axis cs: 9.4, -12)
		        node[midway, below=15pt, font=\huge] {k=4};
		}, 
        title={\huge Vocab size = 4}
        ]
		\addplot [draw=none, fill=mona_lisa]
		coordinates {
		    (1, 60.00) (2, 30.00) (3, 20.0) (4, 30.00) (5, 0) (6, 0.0) (7, 10.00) (8, 0.00) (9, 0)
		}; 
		\addplot [draw=none, fill=flamingo]
		coordinates {
		    (1, 30.00) (2, 10) (3, 10) (4, 0) (5, 0) (6, 0) (7, 10.00) (8, 0.00) (9, 0)
		}; 
        \addplot [draw=none, fill=shakespeare]
		coordinates {
		    (1, 40.00) (2, 10) (3, 0) (4, 20) (5, 0) (6, 0) (7, 20.00) (8, 0.00) (9, 0)
		}; 
		\addplot [draw=none, fill=sunset_orange]
		coordinates {
		    (1, 60) (2, 50) (3, 40) (4, 40) (5, 20) (6, 10) (7, 40) (8, 0) (9, 0)
		}; 
		\addplot[draw=none, fill=sushi] coordinates {
		    (1, 80) (2, 60) (3, 80) (4, 50) (5, 10) (6, 10) (7, 10) (8, 0.00) (9, 0)
		}; 

	\end{groupplot}

\end{tikzpicture} }

    \caption{ISL results for five models: Llama3.3-70b, Llama3.1-405b, GPT-4o, DeepSeek-V3, o1-mini}
    \label{fig:ISL_figure}
\end{figure*}
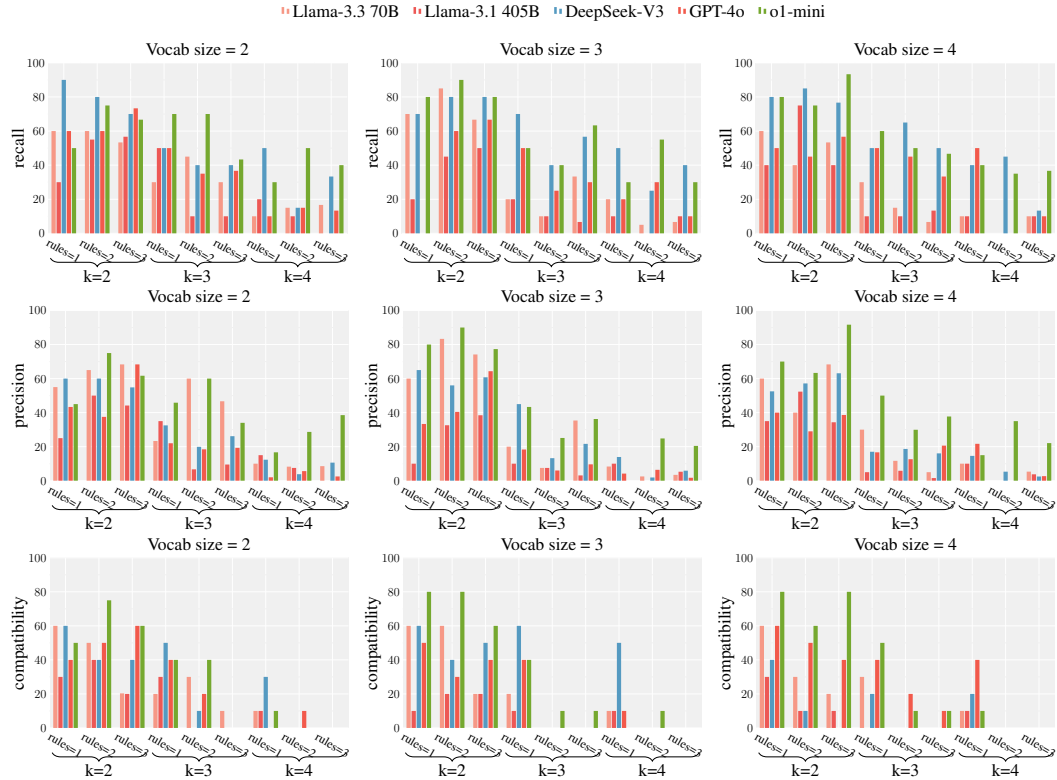

Figure~\ref{fig:ISL_figure} shows how five models (Llama3.3-70B, Llama3.1-405B, GPT-4o, DeepSeek-V3, and o1-mini) perform on various ISL tasks, organized by three key parameters: the context window size $k$, the vocabulary size $|\Sigma|$, and the minimal number of rules required to describe the function. The top row of panels presents recall, the middle row presents precision, and the bottom row presents compatibility. We did not include o3-mini here because its performance is way stronger than all other five models and thereby in order to show the impact of various factors, we omit this model for better visual clarity. Based on these figures, the impact of $k$, $|\Sigma|$, and number of rules become very clear: (1) increasing $k$ markedly reduces recall, precision, and compatibility (2) $|\Sigma|$ does not impact the performance much (3) the number of minimal rules can substantially affect compatibility with large $k$ and $|\Sigma|$. 

\paragraph{Impact of $k$.} Across all models, moving from $k=2$ to $k=4$ markedly reduces recall, precision, and compatibility. This trend underscores how increasing the context window increases the complexity of the underlying ISL functions and making it more challenging for current LLMs to learn the correct transformations. Longer look-ahead requires the model to track additional input context, which can overload its capacity to induce reliable rules.

\paragraph{Impact of $|\Sigma|$.} In contrast, enlarging the vocabulary from $|\Sigma| = 2$ to $|\Sigma| = 4$ does not consistently degrade performance to the same degree as increasing $k$. While some models exhibit slight declines in recall or precision with a larger alphabet, these effects are neither as uniform nor as pronounced as those induced by a bigger Markov window. This finding suggests that the breadth of symbol variation matters less than the depth of sequential dependencies.

\paragraph{Impact of the Number of Rules.} Notably, the number of minimal rules can substantially affect compatibility. When $k=2$ and $|\Sigma| = 2$, a comparatively small search space, changing the number of rules does not drastically alter compatibility. However, under more demanding scenarios where $k\in\{3, 4\}$, the data indicate that adding rules can cause compatibility to plummet. In many cases, having just one rule still yields nontrivial compatibility, whereas introducing a second or third rule often overwhelms the models, resulting in compatibility scores near 0. 

\paragraph{Impact of Examples} We further examined whether few-shot prompting could enhance model performance. In our experiments with Llama-3.3 70B, we varied the number of in-context examples (1-shot, 2-shot, and 3-shot) to determine their effect on the model's ability to induce the correct function representation. The results indicate that when both the vocabulary size and the context window $k$ are small, adding more examples improves performance across the evaluated metrics. However, as the complexity increases—with larger values of $k$ and vocabulary sizes—the benefits of additional few-shot examples become negligible. These findings suggest that while few-shot learning is beneficial for simpler settings, its efficacy diminishes in more complex inductive tasks. Experiment results are presented in Tables \ref{tab:few_shot_ISL}, \ref{tab:few_shot_LOSL}, and \ref{tab:few_shot_ROSL}.

\pgfplotsset{
axis background/.style={fill=gallery!62},
grid=both,
  xtick pos=left,
  ytick pos=left,
  tick style={
    major grid style={style=white, line width=0.5pt},
    minor grid style=gallery!62,
    draw=none
    },
  minor tick num=1,
  ymajorgrids,
	major grid style={draw=white},
	y axis line style={opacity=0},
	tickwidth=0pt,
}


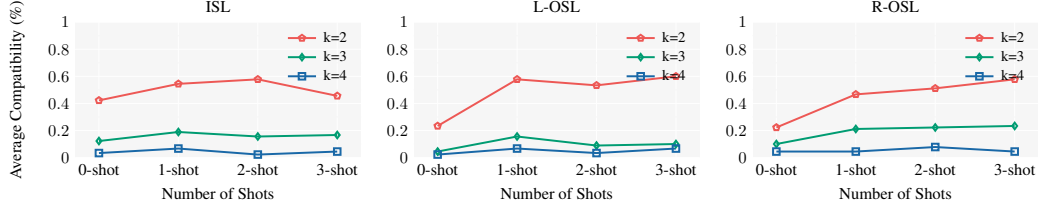
\begin{figure*}[!ht]
\centering
    \begin{tikzpicture} [scale=0.62]
	\begin{groupplot}[
	    group style={group size=3 by 1,
	        horizontal sep = 32pt,
	        }, 
	width=0.55\linewidth,
	height=0.32\linewidth,
        ymin = 0,
        ymax = 1,
	xlabel=Number of Shots,
        xticklabels={0-shot, 1-shot, 2-shot, 3-shot},
        xtick={1,2,3,4,5},
        ymajorgrids,
        major grid style={draw=white},
        y axis line style={opacity=0},
        tickwidth=0pt,
        yticklabel style={
        /pgf/number format/fixed,
        /pgf/number format/precision=5,
        },
        legend style = {
		  font=\small,
          draw=none, 
          fill=none,
          column sep = 2pt, 
          /tikz/every even column/.append style={column sep=5mm},
          legend columns = 1, 
          },
        scaled y ticks=false,
        every axis plot/.append style={mark size=2, very thick},
	    ]
	\nextgroupplot[
		title=ISL, 
            title style={at={(0.5,0.96)}, anchor=south},
            ylabel=Average Compatibility (\%),
		]
		
	\addplot[color=flamingo,mark=pentagon] coordinates {
          (1, 0.4222)
          (2, 0.5444444444444445)
          (3, 0.5777777777777778)
          (4, 0.4555555555555555)
        }; \addlegendentry{k=2}
         \addplot[color=free_speech_aquamarine,mark=diamond] coordinates {
          (1, 0.1222)
          (2, 0.18888888888888888)
          (3, 0.15555555555555556)
          (4, 0.16666666666666666)
        }; \addlegendentry{k=3}
        \addplot[color=matisse,mark=square] coordinates {
          (1, 0.0333)
          (2, 0.06666666666666668)
          (3, 0.022222222222222223)
          (4, 0.044444444444444446)
        };\addlegendentry{k=4}
        
        \nextgroupplot[
        title=L-OSL,
        title style={at={(0.5,0.96)}, anchor=south}, ]
		\addplot[color=flamingo,mark=pentagon] coordinates {
          (1, 0.23333333333333328)
          (2, 0.5777777777777778)
          (3, 0.5333333333333332)
          (4, 0.6000000000000001)
        }; \addlegendentry{k=2}
         \addplot[color=free_speech_aquamarine,mark=diamond] coordinates {
          (1, 0.044444444444444446)
          (2, 0.15555555555555556)
          (3, 0.08888888888888888)
          (4, 0.1)
        }; \addlegendentry{k=3}
        \addplot[color=matisse,mark=square] coordinates {
          (1, 0.022222222222222223)
          (2, 0.06666666666666668)
          (3, 0.03333333333333334)
          (4, 0.06666666666666668)
        };\addlegendentry{k=4}
        
        \nextgroupplot[title=R-OSL, 
        title style={at={(0.5,0.96)}, anchor=south}, ]
	\addplot[color=flamingo,mark=pentagon] coordinates {
          (1, 0.2222222222222222)
          (2, 0.46666666666666656)
          (3, 0.5111111111111111)
          (4, 0.5777777777777778)
        }; \addlegendentry{k=2}
         \addplot[color=free_speech_aquamarine,mark=diamond] coordinates {
          (1, 0.10000000000000002)
          (2, 0.21111111111111114)
          (3, 0.2222222222222222)
          (4, 0.23333333333333334)
        }; \addlegendentry{k=3}
        \addplot[color=matisse,mark=square] coordinates {
          (1, 0.044444444444444446)
          (2, 0.044444444444444446)
          (3, 0.07777777777777778)
          (4, 0.044444444444444446)
        };\addlegendentry{k=4}

	\end{groupplot}

\end{tikzpicture}
\vspace{-10pt}
    \caption{Impact of Few-shot Examples under different $k$.}
\vspace{-10pt}
    \label{fig:fewshot}
\end{figure*}

\paragraph{Robustness} We assess the stability of inductive reasoning by varying the number of input–output pairs provided to the model. The x-axis represent $\frac{|\mathcal{D}|}{|S|}$ where $S$ is the minimal set of examples needed to guarantee learnability of the underlying function. The hypothesis is that if the model were performing genuine logical or inductive reasoning, we would expect performance to remain stable or even improve as more data points become available, since these points should further clarify the underlying function. Figure~\ref{fig:sample_size_impact} illustrates how average compatibility decreases steeply as the number of provided input–output examples increases. This drop suggests that the LLM's reasoning process is not robustly inductive: rather than refining its hypothesis with additional data, the model appears to become confused or overwhelmed, leading to poorer overall performance. Consequently, these findings highlight the limited robustness of current LLMs' inductive reasoning, particularly in scenarios where increasing the available data should theoretically facilitate, rather than hinder, function inference.

\pgfplotsset{
axis background/.style={fill=gallery!62},
grid=both,
  xtick pos=left,
  ytick pos=left,
  tick style={
    major grid style={style=white, line width=0.5pt},
    minor grid style=gallery!62,
    draw=none
    },
  minor tick num=1,
  ymajorgrids,
	major grid style={draw=white},
	y axis line style={opacity=0},
	tickwidth=0pt,
}


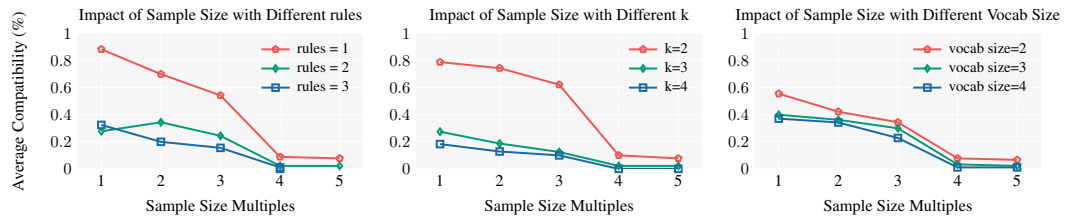
\begin{figure*}
\centering
    \begin{tikzpicture} [scale=0.62]
	\begin{groupplot}[
	    group style={group size=3 by 1,
	        horizontal sep = 32pt,
	        }, 
	width=0.55\linewidth,
	height=0.32\linewidth,
        ymin = 0,
        ymax = 1,
	xlabel=Sample Size Multiples,
        xticklabels={1, 2, 3, 4, 5},
        xtick={1,2,3,4,5,6},
        ymajorgrids,
        major grid style={draw=white},
        y axis line style={opacity=0},
        tickwidth=0pt,
        yticklabel style={
        /pgf/number format/fixed,
        /pgf/number format/precision=5,
        },
        legend style = {
		  font=\small,
          draw=none, 
          fill=none,
          column sep = 2pt, 
          /tikz/every even column/.append style={column sep=5mm},
          legend columns = 1, 
          },
        scaled y ticks=false,
        every axis plot/.append style={mark size=2, very thick},
	    ]
	\nextgroupplot[
		title=Impact of Sample Size with Different rules, 
            title style={at={(0.5,0.96)}, anchor=south},
            ylabel=Average Compatibility (\%),
		]
	\addplot[color=flamingo,mark=pentagon] coordinates {
          (1, 0.8833333333333333)
          (2, 0.6999999999999998)
          (3, 0.5428571428571428)
          (4, 0.08888888888888889)
          (5, 0.07777777777777778)
        }; \addlegendentry{rules = 1}
         \addplot[color=free_speech_aquamarine,mark=diamond] coordinates {
          (1, 0.2777777777777778)
          (2, 0.34444444444444444)
          (3, 0.24444444444444446)
          (4, 0.022222222222222223)
          (5, 0.022222222222222223)
        }; \addlegendentry{rules = 2}
        \addplot[color=matisse,mark=square] coordinates {
          (1, 0.32500000000000007)
          (2, 0.20000000000000004)
          (3, 0.15555555555555556)
          (4, 0.011111111111111112)
          (4, 0.0)
        };\addlegendentry{rules = 3}
        
        \nextgroupplot[
        title=Impact of Sample Size with Different k,
        title style={at={(0.5,0.96)}, anchor=south}, ]
		\addplot[color=flamingo,mark=pentagon] coordinates {
          (1, 0.788888888888889)
          (2, 0.7444444444444445)
          (3, 0.6222222222222222)
          (4, 0.10000000000000002)
          (5, 0.07777777777777778)
        }; \addlegendentry{k=2}
         \addplot[color=free_speech_aquamarine,mark=diamond] coordinates {
          (1, 0.275)
          (2, 0.1875)
          (3, 0.125)
          (4, 0.022222222222222223)
          (5, 0.022222222222222223)
        }; \addlegendentry{k=3}
        \addplot[color=matisse,mark=square] coordinates {
          (1, 0.18333333333333335)
          (2, 0.12857142857142856)
          (3, 0.1)
          (4, 0.0)
          (5, 0.0)
        };\addlegendentry{k=4}
        \nextgroupplot[title=Impact of Sample Size with Different Vocab Size, 
        title style={at={(0.5,0.96)}, anchor=south}, ]
	\addplot[color=flamingo,mark=pentagon] coordinates {
          (1, 0.5555555555555556)
          (2, 0.4222222222222222)
          (3, 0.34444444444444444)
          (4, 0.07777777777777778)
          (5, 0.06666666666666668)
        }; \addlegendentry{vocab size=2}
         \addplot[color=free_speech_aquamarine,mark=diamond] coordinates {
          (1, 0.39999999999999997)
          (2, 0.36250000000000004)
          (3, 0.30000000000000004)
          (4, 0.03333333333333334)
          (5, 0.022222222222222223)
        }; \addlegendentry{vocab size=3}
        \addplot[color=matisse,mark=square] coordinates {
          (1, 0.37142857142857144)
          (2, 0.34285714285714286)
          (3, 0.2285714285714286)
          (4, 0.011111111111111112)
          (5, 0.011111111111111112)
        };\addlegendentry{vocab size=4}

	\end{groupplot}

\end{tikzpicture}
\vspace{-10pt}
    \caption{Impact of Sample Size with Different Sample Size Multiples.}
\vspace{-10pt}
    \label{fig:sample_size_impact}
\end{figure*}

\pgfplotsset{
axis background/.style={fill=gallery!62},
grid=both,
  xtick pos=left,
  ytick pos=left,
  tick style={
    major grid style={style=white, line width=0.5pt},
    minor grid style=gallery!62,
    draw=none
    },
  minor tick num=1,
  ymajorgrids,
	major grid style={draw=white},
	y axis line style={opacity=0},
	tickwidth=0pt,
}


\begin{figure*}
\centering
    \begin{tikzpicture} [scale=0.62]
	\begin{groupplot}[
	    group style={group size=3 by 1,
	        horizontal sep = 32pt,
	        }, 
	width=0.55\linewidth,
	height=0.32\linewidth,
        ymin = 0,
        ymax = 1,
	xlabel=Sample Size Multiples,
        xticklabels={1, 2, 3, 4, 5},
        xtick={1,2,3,4,5,6},
        ymajorgrids,
        major grid style={draw=white},
        y axis line style={opacity=0},
        tickwidth=0pt,
        yticklabel style={
        /pgf/number format/fixed,
        /pgf/number format/precision=5,
        },
        legend style = {
		  font=\small,
          draw=none, 
          fill=none,
          column sep = 2pt, 
          /tikz/every even column/.append style={column sep=5mm},
          legend columns = 1, 
          },
        scaled y ticks=false,
        every axis plot/.append style={mark size=2, very thick},
	    ]
	\nextgroupplot[
		title=Impact of Context Length with Different rules, 
            title style={at={(0.5,0.96)}, anchor=south},
            ylabel=Average Compatibility (\%),
		]
	\addplot[color=flamingo,mark=pentagon] coordinates {
          (1, 0.8833333333333333)
          (2, 0.4833333333333334)
          (3, 0.44285714285714295)
          (4, 0.3111111111111111)
          (5, 0.34444444444444444)
        }; \addlegendentry{rules = 1}
         \addplot[color=free_speech_aquamarine,mark=diamond] coordinates {
          (1, 0.2777777777777778)
          (2, 0.2)
          (3, 0.17777777777777776)
          (4, 0.2222222222222222)
          (5, 0.18888888888888888)
        }; \addlegendentry{rules = 2}
        \addplot[color=matisse,mark=square] coordinates {
          (1, 0.32500000000000007)
          (2, 0.07777777777777778)
          (3, 0.07777777777777778)
          (4, 0.05555555555555555)
          (5, 0.06666666666666667)
        };\addlegendentry{rules = 3}
        
        \nextgroupplot[
        title=Impact of Context Length with Different k,
        title style={at={(0.5,0.96)}, anchor=south}, ]
		\addplot[color=flamingo,mark=pentagon] coordinates {
          (1, 0.788888888888889)
          (2, 0.4555555555555555)
          (3, 0.43333333333333335)
          (4, 0.4)
          (5, 0.4111111111111111)
        }; \addlegendentry{k=2}
         \addplot[color=free_speech_aquamarine,mark=diamond] coordinates {
          (1, 0.275)
          (2, 0.15)
          (3, 0.15000000000000002)
          (4, 0.16666666666666666)
          (5, 0.12222222222222223)
        }; \addlegendentry{k=3}
        \addplot[color=matisse,mark=square] coordinates {
          (1, 0.18333333333333335)
          (2, 0.014285714285714287)
          (3, 0.0375)
          (4, 0.022222222222222223)
          (5, 0.06666666666666667)
        };\addlegendentry{k=4}
        \nextgroupplot[title=Impact of Context Length with Different Vocab Size, 
        title style={at={(0.5,0.96)}, anchor=south}, ]
	\addplot[color=flamingo,mark=pentagon] coordinates {
          (1, 0.5555555555555556)
          (2, 0.3111111111111111)
          (3, 0.25555555555555554)
          (4, 0.23333333333333334)
          (5, 0.2777777777777778)
        }; \addlegendentry{vocab size=2}
         \addplot[color=free_speech_aquamarine,mark=diamond] coordinates {
          (1, 0.39999999999999997)
          (2, 0.25)
          (3, 0.2777777777777778)
          (4, 0.2666666666666667)
          (5, 0.24444444444444446)
        }; \addlegendentry{vocab size=3}
        \addplot[color=matisse,mark=square] coordinates {
          (1, 0.37142857142857144)
          (2, 0.08571428571428572)
          (3, 0.08571428571428573)
          (4, 0.08888888888888889)
          (5, 0.07777777777777778)
        };\addlegendentry{vocab size=4}

	\end{groupplot}

\end{tikzpicture}
\vspace{-10pt}
    \caption{Impact of Context Length with Different Sample Size Multiples.}
\vspace{-10pt}
    \label{fig:context_length_impact}
\end{figure*}
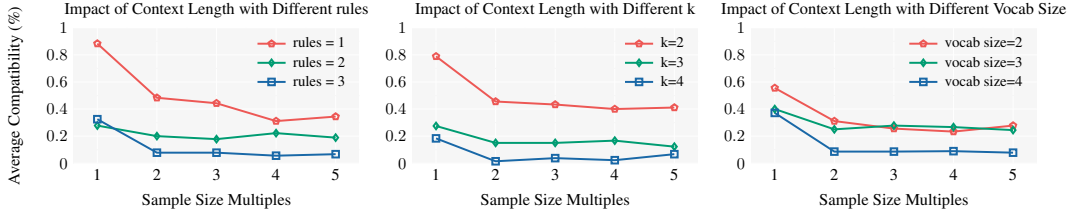

Moreover, to isolate the influence of context length from the effect of adding genuinely new data, we conduct an additional experiment in which we simply extending the context size by \emph{repeating} the minimal characteristic sample without introducing novel input–output pairs. Comparing Figures~\ref{fig:sample_size_impact} and \ref{fig:context_length_impact} reveals that while compatibility does diminish with increased context length (\emph{e.g.}, at a multiple of 2), the decline is relatively small when scaling further to multiples of 3, 4, or 5. By contrast, when truly new datapoints are added (and not just repeated), compatibility plummets nearly to zero for multiples of 4 and 5. These results confirm that the primary driver of performance degradation is the inclusion of additional, distinct datapoints rather than simply lengthening the context.

\paragraph{Error Type Analysis} We further examined the specific types of errors made by LLMs when their predicted functions failed to match the ground-truth dataset. At a high level, we distinguish between \emph{missing rules} (leading to low recall) and \emph{wrong rules} (leading to low precision).

\textbf{Missing Rules:}
These refer to ground-truth rules that do not appear in the model's predicted rule set. We classify missing rules into three subtypes: 
\begin{enumerate} \item \emph{Too General.} Although a certain ground-truth rule $r:c\circ u\to v$ was missed, there exists a corresponding predicted rule $r': c'\circ u'\to v'$ that over-generalizes. Specifically, the condition $c'$ is a \emph{proper suffix} of $c$, causing $r'$ to apply more broadly than intended.
\item \emph{Too Specific.} The opposite of the above: a predicted rule condition $c'$ is a proper \emph{extension} of $c$, thus applying too narrowly and failing to match some instances that should have been captured by the ground-truth rule.
\item \emph{Completely Missed.} No predicted rule over-generalizes or under-generalizes the ground-truth rule; in other words, this pattern is simply absent from the predicted rule set altogether. \end{enumerate}

\textbf{Wrong Rules:}  
These refer to rules present in the model's predicted set that do not exist in the ground truth. We categorize such rules into four types:
\begin{enumerate}
    \item \emph{Too General.} The rule \(r': c' \circ u' \to v'\) is overly broad, applying in contexts where the ground truth does not. This typically arises when \(c'\) is a proper suffix of some genuine condition \(c\) and thus fails to capture necessary constraints.  
    \item \emph{Too Specific.} The rule narrowly addresses only a subset of the intended patterns (e.g., by employing a condition \(c'\) that is an extension of the legitimate condition \(c\)), thereby missing broader contexts that should have matched.  
    \item \emph{Correct Condition but Wrong Transformation.} Here, the predicted rule accurately identifies the correct condition \(c'\) and target input character \(u'\), but the transformation \(v'\) is incorrect.  
    \item \emph{Completely Wrong.} None of the above criteria apply: the rule's condition and transformation are both inconsistent with the ground truth, indicating a fundamental misunderstanding.
\end{enumerate}

\begin{figure*}[!ht]
    \centering
    \includegraphics[scale=0.25]{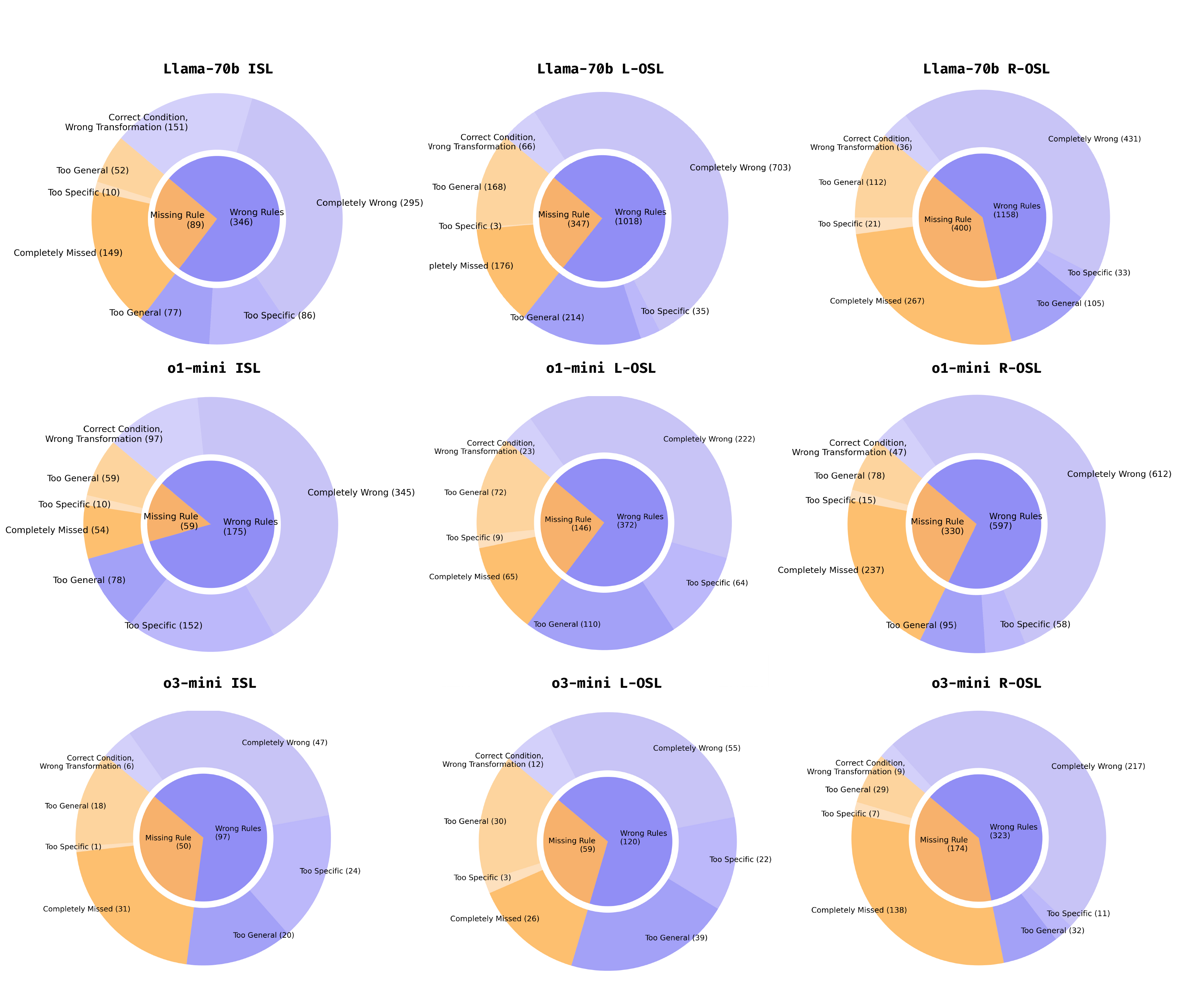}
    \caption{Error Type Analysis}
    \label{fig:error_analysis}
\end{figure*}

We present a breakdown of error types for three models in Figure\ref{fig:error_analysis}: Llama3.3-70B, o1-mini, and o3-mini. Among \emph{missing rules}, the most common issue is \textbf{completely missed}, where the model fails to identify the relevant pattern at all. The second most frequent error is \textbf{too general}, suggesting that the predicted condition is shorter than needed, thereby overgeneralizing the intended behavior. In contrast, \textbf{too specific} errors in this category are relatively rare. Among \emph{wrong rules}, the majority are \textbf{completely wrong}, followed by a notable fraction of \textbf{too general}. Although there is also a non-negligible number of \textbf{too specific} errors, these tend to occur when a single ground-truth rule (e.g., $ab\circ c\to b$ is replaced by multiple subcases (\emph{e.g.}, $aab\circ c\to b, bab\circ c\to b, cab\circ c\to b$, indicating the model has uncovered individual instances but failed to unify them into a concise representation. Finally, \textbf{correct condition but wrong transformation} occurs relatively infrequently, implying that once the model infers the correct condition pattern, it typically produces the correct transformation.

\subsection{Summary of findings}
Overall, our experiments reveal four main insights into the inductive reasoning performance of current LLMs:
\begin{itemize}
    \item Context window size $k$ dominates complexity: Increasing $k$ from 2 to 4 significantly degrades recall, precision, and compatibility, underscoring how longer look-ahead windows intensify the complexity of ISL functions. 
    \item Number of Rules increases difficulty under large hypothesis space: The number of minimal rules required can drastically lower compatibility in more challenging settings with large $k$ adn $|\Sigma|$, indicating that managing multiple interacting rules overwhelms many models.
    \item Few-shot examples do not help much: Few-shot examples help in simpler configurations but yield diminishing returns as $k$ adn $|\Sigma|$ grows—suggesting that, past a certain complexity threshold, additional examples do not compensate for the model's limited inductive capacity.
    \item Current LLMs are very unrobust: Providing more novel data should theoretically clarify function patterns, yet performance often plummets, reflecting a fragility in inductive reasoning.
    \item Error analysis shows that missing rules are most frequently ``completely missed'' or ``too general,'' while wrong rules often end up ``completely wrong'' or again ``too general.'' Only a small fraction are ``too specific'' or feature a ``correct condition but wrong transformation,'' indicating that once models identify the right condition, they typically produce the correct transformation.
\end{itemize}

Taken together, these findings highlight fundamental limits in current LLMs' inductive reasoning. Even state-of-the-art models often fail as complexity grows, or when confronted with more data than their inductive mechanisms appear able to systematically absorb.

\section{Leaderboard based on InductionBench}
To facilitate straightforward comparisons among different LLMs, we introduce a two-part benchmark leaderboard: a \emph{standard leaderboard} and an \emph{exploration leaderboard}. The \emph{standard leaderboard} is based completely on the three function classes we talked about, and this leaderboard simply presents an aggregated score to directly reflect LLM's performance. The \emph{exploration leaderboard} includes a slightly new design of function class and we will present the motivation and details below. 

\subsection{Standard Leaderboard}
The standard leaderboard consists of 1,080 questions spanning three classes of deterministic regular functions: \emph{ISL}, \emph{L-OSL}, and \emph{R-OSL} in equal proportion. Specifically, it includes: \begin{itemize}[leftmargin=*, itemsep=1pt] \item 360 ISL questions, \item 360 L-OSL questions, \item 360 R-OSL questions. \end{itemize}

Within each function class, we have settings for $k\in\{2, 3, 4\}, |\Sigma|\in\{5, 6, 7, 8\}$, and number of rules $\in\{3, 4, 5\}$. Each unique parameter combination has 10 data points, totaling 360 points per function class. The performance metrics \emph{recall}, \emph{precision}, and \emph{compatibility} are computed on a per-setting basis. We then form an overall \emph{weighted average} to account for variations in function-space size: 

\begin{definition}
For a given setting characterized by $(k, |\Sigma|, r)$, the weight $w$ is defined as $\frac{|\Sigma|^k}{\sum\limits_{k'=2}^{k'=4}\sum\limits_{s=5}^{s=8}s^{k'}}$, where $k$ is the Markov window, $|\Sigma|$ is the alphabet size, and $r$ is the minimal rule count.
\end{definition}

For each function class (ISL, L-OSL, R-OSL), we compute a weighted recall, precision, and compatibility according to the above scheme and then take the average of these three scores to produce the final leaderboard score for that class. The overall score across all three classes is the average of those class-wise scores.

Table~\ref{tab:leaderboard} summarizes current leaderboard results for several representative models. Notably, even o3-mini achieves only a 
5.69\% compatibility score, largely because none of the models succeed on tasks where $k=4$. Since those high-complexity settings receive substantially larger weights than cases where $k\in\{2, 3\}$, they disproportionately reduce the overall average.

\begin{table}[!ht]
    \centering
    \begin{tabular}{lccc}
    \toprule
    model & average recall & average precision & average compatibility \\
    \midrule
    Llama-3.1 8b & 0.00&	0.00&	0.00\\
    Qwen2.5-Coder-32B-Instruct	&7.26	&0.66	&0.03\\
    Llama-3.3-70b	&6.55	&5.76&	0.12\\
    DeepSeek-R1-Distill-Llama-70B	&3.84	&5.14	&0.78\\
    o3-mini	&28.93	&43.12	&5.69\\
    \bottomrule
    \end{tabular}
    \caption{Leaderboard Result}
    \label{tab:leaderboard}
\end{table}

To balance the influence of different complexity settings, we additionally report an alternative evaluation metric that replaces each original weight with its logarithm. This approach dampens the dominance of $k=4$ scenarios, yielding a more even distribution of weights across the benchmark’s parameter space.

\begin{table}[!ht]
    \centering
    \begin{tabular}{lccc}
    \toprule
    model & average recall & average precision & average compatibility \\
    \midrule
    Llama-3.1 8b&	0.00	&0.00	&0.00\\
    Qwen2.5-Coder-32B-Instruct	&7.48	&6.60	&0.48\\
    Llama-3.3-70b	&8.71	&7.50	&0.87\\
    DeepSeek-R1-Distill-Llama-70B	&23.17&	24.66&	8.63\\
    o3-mini	&57.58	&63.89&33.93\\
    \bottomrule
    \end{tabular}
    \caption{Leaderboard Result with Log Weight}
    \label{tab:leaderboard_log}
\end{table}

\subsection{Exploration Leaderboard}
A key concern in using subregular function classes (\emph{e.g.}, ISL, L-OSL, R-OSL) is that polynomial-time learning algorithms already exist for these classes, potentially allowing a trivial ``hack'' to achieve artificially high performance. Though we advocate not using the provbly correct algorithm for task solving so that we can genuinely evaluate LLM's inductive reasoning ability, to make sure, we introduce an \emph{exploration leaderboard} that focuses on \emph{Input-Output Strictly Local (IOSL)} functions: a more speculative class for which no known algorithm can reliably learn the entire function from finite data in finite time.

\paragraph{Rationale.} Since IOSL lacks a proven polynomial-time learning procedure, successful performance here would more credibly reflect genuine inductive reasoning rather than the application of a known ``shortcut'' algorithm. Furthermore, IOSL functions have not been deeply studied in the literature, offering an opportunity to see whether LLMs can advance this open research area.

This is the definition of IOSL:
\begin{definition}[IOSL]
A function f is IOSL if there is a $k$ such that for all $u_1, u_2\in\Sigma^*$, if $\textsc{Suff}^{k-1}(u_1) = \textsc{Suff}^{k-1}(u_2)$ and $\textsc{Suff}^{k-1}(f(u_1)) = \textsc{Suff}^{k-1}(f(u_2))$, then $\textsc{tails}_f(u_1) = \textsc{tails}_f(u_2)$.
\end{definition}

In essence, this condition requires the model to distinguish between input-based and output-based Markovian triggers, making the learned transformation highly non-trivial if no pre-existing algorithm is used.

\paragraph{Leaderboard Setup.} The IOSL-based leaderboard contains 1{,}080 datapoints, mirroring the standard leaderboard in overall structure: $k\in\{2, 3, 4\}, |\Sigma|\in\{5, 6, 7, 8\}$, number of rules $\in\{3, 4, 5\}$. For each setting, there are 30 datapoints per setting (for equivalence to the standard leaderboard’s size).

Since IOSL is not known to admit a finite-characteristic sample or minimal representation in the same sense as the deterministic classes, we introduce two adaptations for evaluation:

\begin{enumerate}[leftmargin=*, itemsep=1pt] \item \textbf{Sample Size.} We arbitrarily fix the sample size at $2*|\Sigma|^k$, as no characteristic sample is theoretically guaranteed.

\item \textbf{Evaluation Metrics.} We focus primarily on \emph{compatibility}, as recall and precision hinge on the assumption of a unique minimal-length description, which may not exist for IOSL. If a model's generated rule set is compatible with the data, we then check whether its description length is shorter, identical, or longer than our function's reference length. A longer description indicates a definite failure to produce a minimal representation; shorter or equal does not guarantee minimality, but it at least suggests the model avoids obvious redundancy. \end{enumerate}

By presenting both a standard leaderboard (subregular classes with known learnability) and an exploration leaderboard (IOSL with no established finite-data algorithm), we offer a balanced view: models can demonstrate success in theoretically well-understood tasks while also exploring novel, under-constrained function classes—thereby reducing the concern that high performance might merely reflect an existing ``hack.''

\section{Conclusion} 

In this work, we introduced a systematic benchmark for assessing the inductive reasoning capabilities of LLMs, leveraging both well-studied subregular function classes (ISL, L-OSL, and R-OSL) and a more exploratory class (IOSL) for which no known polynomial-time learning algorithm exists. By controlling parameters such as the Markov window size $k$, the vocabulary size $|\Sigma|$, and the minimal number of rules, we offered precise yet flexible tasks capable of probing a model’s capacity to infer general transformations from limited data. Our findings revealed several significant challenges for current LLMs—especially when required to track deeper dependencies or manage larger search spaces—and underscored the fragility of their inductive reasoning under increased context or novel data.

Through experiments measuring recall, precision, and compatibility, we demonstrated that factors like the Markov window size $k$ and the number of rules more profoundly degrade performance than an expanded alphabet. Moreover, while few-shot prompting showed promise in simpler scenarios, its benefits quickly plateaued in more complex contexts. An error analysis further highlighted how many rules go completely missing or become overgeneralized under stringent settings, indicating that LLMs often fail to synthesize key patterns comprehensively.

We also proposed an exploration leaderboard targeting IOSL functions, a class beyond established theoretical learnability, to address concerns that performance gains might stem from known polynomial-time algorithms rather than genuine inductive reasoning. This complementary evaluation opens avenues for research on less tractable classes and poses a more authentic test of generalization and adaptability.

Overall, our results highlight the need for more robust inductive reasoning strategies within current LLM architectures. We hope that our benchmark will help catalyze progress in both theoretical understanding and practical innovations around LLMs' inductive capabilities.

\section*{Limitations}

While our benchmark offers a rigorous, theoretically grounded approach to evaluating inductive reasoning in LLMs, current paper is subject to two notable constraints:

\paragraph{Synthetic Rather Than Real-World Data.} All tasks and evaluations rely on functions generated from carefully controlled parameters rather than naturally occurring texts or real-world datasets. Although this design enables precise measurement of inductive capabilities, it may not fully capture the complexity of practical language use, where ambiguous contexts, noisy inputs, and domain-specific factors can further challenge inference.

\paragraph{Restricted Access to the o1 Model.} Our investigation into the o1 family of models is hindered by limited availability and computational resources. As a result, certain aspects of o1’s inductive behavior may remain unexamined, and a more exhaustive exploration of variations or fine-tuning strategies for o1 could further illuminate its performance.

\bibliography{iclr2025_conference}
\bibliographystyle{iclr2025_conference}

\appendix
\section{Appendix}
Full results on ISL, OSL, and few-shot experiments are presented here.

\clearpage
\begin{table*}[t]
    \centering
    \renewcommand{\arraystretch}{1.1}
    \resizebox{\textwidth}{!}{
    \begin{tabular}{l c ccc ccc ccc}
        \toprule
          \multirow{2}{*}{\bf Models}& \multirow{2}{*}{\bf Settings}& \multicolumn{3}{c}{\bf k = 2} & \multicolumn{3}{c}{\bf k = 3} & \multicolumn{3}{c}{\bf k = 4}\\
          \cmidrule(lr){3-5} \cmidrule(lr){6-8}  \cmidrule(lr){9-11}
          & & recall  & precision & compatibility & recall  & precision & compatibility & recall  & precision & compatibility \\
         \midrule 
         \multicolumn{11}{c}{\textbf{vocab size = 2}} \\
         \midrule
           \multirow{3}{*}{Llama-3.3 70B} & rules = 1 & 60.00 & 55.00 & 60.00 & 30.00 & 23.33 & 20.00 & 10.00 & 10.00 & 10.00\\
           & rules = 2 & 60.00 & 65.00 & 50.00 & 45.00 & 60.00 & 30.00 & 15.00 & 8.25 & 0.00\\
           & \cellcolor{SeaGreen3!15}rules = 3 & \cellcolor{SeaGreen3!15}53.33 & \cellcolor{SeaGreen3!15}68.33 & \cellcolor{SeaGreen3!15}20.00 & \cellcolor{SeaGreen3!15}30.00 & \cellcolor{SeaGreen3!15}46.67 & \cellcolor{SeaGreen3!15}10.00 & \cellcolor{SeaGreen3!15}16.67 & \cellcolor{SeaGreen3!15}8.54 & \cellcolor{SeaGreen3!15}0.00\\
           \hdashline
           \multirow{3}{*}{Llama-3.1 405B} & rules = 1 & 30.00 & 25.00 & 30.00 & 50.00 & 35.00 & 30.00 & 20.00 & 15.00 & 10.00\\
           & rules = 2 & 55.00 & 50.00 & 40.00 & 10.00 & 6.67 & 0.00 & 10.00 & 7.50 & 0.00\\
           & \cellcolor{SeaGreen3!15}rules = 3 & \cellcolor{SeaGreen3!15}56.67 & \cellcolor{SeaGreen3!15}44.17 & \cellcolor{SeaGreen3!15}20.00 & \cellcolor{SeaGreen3!15}10.00 & \cellcolor{SeaGreen3!15}9.50 & \cellcolor{SeaGreen3!15}0.00 & \cellcolor{SeaGreen3!15}0.00 & \cellcolor{SeaGreen3!15}0.00 & \cellcolor{SeaGreen3!15}0.00\\
           \hdashline
           \multirow{3}{*}{DeepSeek-V3} & rules = 1 & 90.00 & 60.00 & 60.00 & 50.00 & 32.50 & 50.00 & 50.00 & 12.33 & 30.00\\
           & rules = 2 & 80.00 & 60.00 & 40.00 & 40.00 & 19.89 & 10.00 & 15.00 & 3.82 & 0.00\\
           & \cellcolor{SeaGreen3!15}rules = 3 & \cellcolor{SeaGreen3!15}70.00 & \cellcolor{SeaGreen3!15}54.83 & \cellcolor{SeaGreen3!15}40.00 & \cellcolor{SeaGreen3!15}40.00 & \cellcolor{SeaGreen3!15}26.15 & \cellcolor{SeaGreen3!15}0.00 & \cellcolor{SeaGreen3!15}33.33 & \cellcolor{SeaGreen3!15}10.61 & \cellcolor{SeaGreen3!15}0.00\\
           \hdashline
           \multirow{3}{*}{GPT-4o} & rules = 1 & 60.00 & 43.33 & 40.00 & 50.00 & 22.00 & 40.00 & 10.00 & 2.00 & 0.00 \\
           & rules = 2 & 60.00 & 37.50 & 50.00 & 35.00 & 18.43 & 20.00 & 15.00 & 5.63 & 10.00 \\
           & \cellcolor{SeaGreen3!15}rules = 3 & \cellcolor{SeaGreen3!15}73.33 & \cellcolor{SeaGreen3!15}68.33 & \cellcolor{SeaGreen3!15}60.00 & \cellcolor{SeaGreen3!15}36.67 & \cellcolor{SeaGreen3!15}19.30 & \cellcolor{SeaGreen3!15}0.00 & \cellcolor{SeaGreen3!15}13.33 & \cellcolor{SeaGreen3!15}2.50 & \cellcolor{SeaGreen3!15}0.00\\
           \hdashline
           \multirow{3}{*}{o1-mini} & rules = 1 & 50.00 & 45.00 & 50.00 & 70.00 & 45.83 & 40.00 & 30.00 & 16.67 & 10.00 \\
           & rules = 2 & 75.00 & 75.00 & 75.00 & 70.00 & 60.00 & 40.00 & 50.00 & 28.67 & 0.00\\
           & \cellcolor{SeaGreen3!15}rules = 3 & \cellcolor{SeaGreen3!15}66.67 & \cellcolor{SeaGreen3!15}61.67 & \cellcolor{SeaGreen3!15}60.00 & \cellcolor{SeaGreen3!15}43.33 & \cellcolor{SeaGreen3!15}34.00 & \cellcolor{SeaGreen3!15}0.00 & \cellcolor{SeaGreen3!15}40.00 & \cellcolor{SeaGreen3!15}38.52 & \cellcolor{SeaGreen3!15}0.00\\
           \hdashline
           \multirow{3}{*}{o3-mini} & rules = 1 & 100.00 & 100.00 & 100.00 & 100.00 & 100.00 & 100.00 & 80.00 & 58.33 & 40.00\\
           & rules = 2 & 90.00 & 90.00 & 90.00 & 90.00 & 90.67 & 90.00 & 85.00 & 80.00 & 50.00\\
           & \cellcolor{SeaGreen3!15}rules = 3 & \cellcolor{SeaGreen3!15}\textbf{100.00} & \cellcolor{SeaGreen3!15}\textbf{97.50} & \cellcolor{SeaGreen3!15}\textbf{100.00} & \cellcolor{SeaGreen3!15}\textbf{83.33} & \cellcolor{SeaGreen3!15}\textbf{71.83} & \cellcolor{SeaGreen3!15}\textbf{60.00} & \cellcolor{SeaGreen3!15}\textbf{70.00} & \cellcolor{SeaGreen3!15}\textbf{63.81} & \cellcolor{SeaGreen3!15}\textbf{20.00}\\
           \midrule 
           \multicolumn{11}{c}{\textbf{vocab size = 3}} \\
         \midrule
           \multirow{3}{*}{Llama-3.3 70B} & rules = 1 & 70.00 & 60.00 & 60.00 & 20.00 & 20.00 & 20.00 & 20.00 & 8.33 & 10.00\\
           & rules = 2 & 85.00 & 83.33 & 60.00 & 10.00 & 7.50 & 0.00 & 5.00 & 2.50 & 0.00\\
           & \cellcolor{SeaGreen3!30}rules = 3 & \cellcolor{SeaGreen3!30}66.67 & \cellcolor{SeaGreen3!30}74.17 & \cellcolor{SeaGreen3!30}20.00 & \cellcolor{SeaGreen3!30}33.33 & \cellcolor{SeaGreen3!30}35.36 & \cellcolor{SeaGreen3!30}0.00 & \cellcolor{SeaGreen3!30}6.67 & \cellcolor{SeaGreen3!30}3.43 & \cellcolor{SeaGreen3!30}0.00 \\
           \hdashline
           \multirow{3}{*}{Llama-3.1 405B} & rules = 1 & 20.00 & 10.00 & 10.00 & 20.00 & 10.00 & 10.00 & 10.00 & 10.00 & 10.00\\
           & rules = 2 & 45.00 & 32.58 & 20.00 & 10.00 & 7.50 & 0.00 & 0.00 & 0.00 & 0.00\\
           & \cellcolor{SeaGreen3!30}rules = 3 & \cellcolor{SeaGreen3!30}50.00 & \cellcolor{SeaGreen3!30}38.45 & \cellcolor{SeaGreen3!30}20.00 & \cellcolor{SeaGreen3!30}6.67 & \cellcolor{SeaGreen3!30}3.10 & \cellcolor{SeaGreen3!30}0.00 & \cellcolor{SeaGreen3!30}10.00 & \cellcolor{SeaGreen3!30}5.27 & \cellcolor{SeaGreen3!30}0.00\\
           \hdashline
           \multirow{3}{*}{DeepSeek-V3} & rules = 1 & 70.00 & 65.00 & 60.00 & 70.00 & 45.00 & 60.00 & 50.00 & 13.93 & 50.00\\
           & rules = 2 & 80.00 & 56.00 & 40.00 & 40.00 & 13.23 & 0.00 & 25.00 & 1.89 & 0.00\\
           & \cellcolor{SeaGreen3!30}rules = 3 & \cellcolor{SeaGreen3!30}80.00 & \cellcolor{SeaGreen3!30}60.76 & \cellcolor{SeaGreen3!30}50.00 & \cellcolor{SeaGreen3!30}56.67 & \cellcolor{SeaGreen3!30}21.64 & \cellcolor{SeaGreen3!30}0.00 & \cellcolor{SeaGreen3!30}40.00 & \cellcolor{SeaGreen3!30}5.88 & \cellcolor{SeaGreen3!30}0.00\\
           \hdashline
           \multirow{3}{*}{GPT-4o} & rules = 1 & 50.00 & 33.33 & 50.00 & 50.00 & 18.33 & 40.00 & 20.00 & 4.17 & 10.00\\
           & rules = 2 & 60.00 & 40.42 & 30.00 & 25.00 & 6.00 & 0.00 & 30.00 & 6.39 & 0.00 \\
           & \cellcolor{SeaGreen3!30}rules = 3 & \cellcolor{SeaGreen3!30}66.67 & \cellcolor{SeaGreen3!30}64.33 & \cellcolor{SeaGreen3!30}40.00 & \cellcolor{SeaGreen3!30}30.00 & \cellcolor{SeaGreen3!30}9.61 & \cellcolor{SeaGreen3!30}0.00 & \cellcolor{SeaGreen3!30}10.00 & \cellcolor{SeaGreen3!30}1.72 & \cellcolor{SeaGreen3!30}0.00\\
           \hdashline
           \multirow{3}{*}{o1-mini} & rules = 1 & 80.00 & 80.00 & 80.00 & 50.00 & 43.33 & 40.00 & 30.00 & 0.00 & 0.00\\
           & rules = 2 & 90.00 & 90.00 & 80.00 & 40.0 & 25.11 & 10.00 & 55.00 & 24.82 & 10.00\\
           & \cellcolor{SeaGreen3!30}rules = 3 & \cellcolor{SeaGreen3!30}80.00 & \cellcolor{SeaGreen3!30}77.33 & \cellcolor{SeaGreen3!30}60.00 & \cellcolor{SeaGreen3!30}63.33 & \cellcolor{SeaGreen3!30}36.25 & \cellcolor{SeaGreen3!30}10.00 & \cellcolor{SeaGreen3!30}30.00 & \cellcolor{SeaGreen3!30}20.44 & \cellcolor{SeaGreen3!30}0.00\\
           \hdashline
           \multirow{3}{*}{o3-mini} & rules = 1 & 100.00 & 100.000 & 100.00 & 100.00 & 95.00 & 90.00 & 90.00 & 78.33 & 70.00\\
           & rules = 2 & 100.00 & 100.000 & 100.00 & 95.00 & 91.67 & 80.00 & 75.00 & 75.00 & 50.00\\
           & \cellcolor{SeaGreen3!30}rules = 3 & \cellcolor{SeaGreen3!30}\textbf{96.67} & \cellcolor{SeaGreen3!30}\textbf{97.50} & \cellcolor{SeaGreen3!30}\textbf{80.00} & \cellcolor{SeaGreen3!30}\textbf{93.33} & \cellcolor{SeaGreen3!30}\textbf{91.67} & \cellcolor{SeaGreen3!30}\textbf{90.00} & \cellcolor{SeaGreen3!30}\textbf{83.33} & \cellcolor{SeaGreen3!30}\textbf{85.17} & \cellcolor{SeaGreen3!30}\textbf{50.00}\\
           \midrule 
           \multicolumn{11}{c}{\textbf{vocab size = 4}} \\
         \midrule
           \multirow{3}{*}{Llama-3.3 70B} & rules = 1 & 60.00 & 60.00 & 60.00 & 30.00 & 30.00 & 30.00 & 10.00 & 10.00 & 10.00\\
           & rules = 2 & 40.00 & 40.00 & 30.00 & 15.00 & 11.67 & 0.00 & 0.00 & 0.00 & 0.00\\
           & \cellcolor{SeaGreen3!50}rules = 3 & \cellcolor{SeaGreen3!50}53.33 & \cellcolor{SeaGreen3!50}68.33 & \cellcolor{SeaGreen3!50}20.00 & \cellcolor{SeaGreen3!50}6.67 & \cellcolor{SeaGreen3!50}5.00 & \cellcolor{SeaGreen3!50}0.00 & \cellcolor{SeaGreen3!50}10.00 & \cellcolor{SeaGreen3!50}5.32 & \cellcolor{SeaGreen3!50}0.00\\
           \hdashline
           \multirow{3}{*}{Llama-3.1 405B} & rules = 1 & 40.00 & 35.00 & 30.00 & 10.00 & 5.00 & 0.00 & 10.00 & 10.00 & 10.00\\
           & rules = 2 & 75.00 & 52.33 & 10.00 & 10.00 & 5.83 & 0.00 & 0.00 & 0.00 & 0.00\\
           & \cellcolor{SeaGreen3!50}rules = 3 & \cellcolor{SeaGreen3!50}40.00 & \cellcolor{SeaGreen3!50}34.33 & \cellcolor{SeaGreen3!50}10.00 & \cellcolor{SeaGreen3!50}13.33 & \cellcolor{SeaGreen3!50}1.54 & \cellcolor{SeaGreen3!50}0.00 & \cellcolor{SeaGreen3!50}10.00 & \cellcolor{SeaGreen3!50}3.75 & \cellcolor{SeaGreen3!50}0.00\\
           \hdashline
           \multirow{3}{*}{DeepSeek-V3} & rules = 1 & 80.00 & 52.50 & 60.00 & 50.00 & 17.00 & 40.00 & 40.00 & 14.58 & 40.00\\
           & rules = 2 & 85.00 & 57.15 & 50.00 & 65.00 & 18.66 & 20.00 & 45.00 & 5.30 & 0.00\\
           & \cellcolor{SeaGreen3!50}rules = 3 & \cellcolor{SeaGreen3!50}76.67 & \cellcolor{SeaGreen3!50}63.12 & \cellcolor{SeaGreen3!50}40.00 & \cellcolor{SeaGreen3!50}50.00 & \cellcolor{SeaGreen3!50}16.05 & \cellcolor{SeaGreen3!50}10.00 & \cellcolor{SeaGreen3!50}13.33 & \cellcolor{SeaGreen3!50}2.46 & \cellcolor{SeaGreen3!50}0.00\\
           \hdashline
           \multirow{3}{*}{GPT-4o} & rules = 1 & 50.00 & 40.00 & 40.00 & 50.00 & 16.67 & 20.00 & 50.00 & 21.67 & 20.00\\
           & rules = 2 & 45.00 & 29.00 & 10.00 & 45.00 & 12.62 & 0.00 & 0.00 & 0.00 & 0.00\\
           & \cellcolor{SeaGreen3!50}rules = 3 & \cellcolor{SeaGreen3!50}56.67 & \cellcolor{SeaGreen3!50}38.62 & \cellcolor{SeaGreen3!50}0.00 & \cellcolor{SeaGreen3!50}33.33 & \cellcolor{SeaGreen3!50}20.60 & \cellcolor{SeaGreen3!50}0.00 & \cellcolor{SeaGreen3!50}10.00 & \cellcolor{SeaGreen3!50}2.67 & \cellcolor{SeaGreen3!50}0.00\\
           \hdashline
           \multirow{3}{*}{o1-mini} & rules = 1 & 80.00 & 70.00 & 80.00 & 60.00 & 50.00 & 50.00 & 40.00 & 15.00 & 10.00\\
           & rules = 2 & 75.00 & 63.33 & 60.00 & 50.00 & 29.93 & 10.00 & 35.00 & 35.00 & 0.00 \\
           & \cellcolor{SeaGreen3!50}rules = 3 & \cellcolor{SeaGreen3!50}93.33 & \cellcolor{SeaGreen3!50}91.67 & \cellcolor{SeaGreen3!50}80.00 & \cellcolor{SeaGreen3!50}46.67 & \cellcolor{SeaGreen3!50}37.72 & \cellcolor{SeaGreen3!50}10.00 & \cellcolor{SeaGreen3!50}36.67 & \cellcolor{SeaGreen3!50}22.09 & \cellcolor{SeaGreen3!50}0.00\\
           \hdashline
           \multirow{3}{*}{o3-mini} & rules = 1 & 100.00 & 100.00 & 100.00 & 100.00 & 95.00 & 100.00 & 60.00 & 60.00 & 60.00\\
           & rules = 2 & 100.00 & 100.00 & 100.00 & 95.00 & 91.67 & 80.00 & 75.00 & 76.67 & 40.00\\
           & \cellcolor{SeaGreen3!50}rules = 3 & \cellcolor{SeaGreen3!50}\textbf{96.67} & \cellcolor{SeaGreen3!50}\textbf{95.00} & \cellcolor{SeaGreen3!50}\textbf{90.00} & \cellcolor{SeaGreen3!50}\textbf{93.33} & \cellcolor{SeaGreen3!50}\textbf{93.33} & \cellcolor{SeaGreen3!50}\textbf{80.00} & \cellcolor{SeaGreen3!50}\textbf{73.33} & \cellcolor{SeaGreen3!50}\textbf{59.58} & \cellcolor{SeaGreen3!50}\textbf{10.00}\\
           \bottomrule
    \end{tabular}
    }
    \caption{Input Strictly Local with sample size = 2}
    \label{tab:ISL_main}
\end{table*}

\begin{table*}[t]
    \centering
    \renewcommand{\arraystretch}{1.1}
    \resizebox{14.5cm}{!}{
    \begin{tabular}{l c ccc ccc ccc}
        \toprule
          \multirow{2}{*}{\bf Models}& \multirow{2}{*}{\bf Settings}& \multicolumn{3}{c}{\bf k = 2} & \multicolumn{3}{c}{\bf k = 3} & \multicolumn{3}{c}{\bf k = 4}\\
          \cmidrule(lr){3-5} \cmidrule(lr){6-8}  \cmidrule(lr){9-11}
          & & recall  & precision & compatibility & recall  & precision & compatibility & recall  & precision & compatibility \\
         \midrule 
         \multicolumn{11}{c}{\textbf{vocab size = 2}} \\
         \midrule
           \multirow{3}{*}{Llama-3.3 70B} & rules = 1 & 50.00 & 45.00 & 50.00 & 0.00 & 0.00 & 0.00 & 0.00 & 0.00 & 0.00\\
           & rules = 2 & 25.00 & 25.00 & 20.00 & 10.00 & 8.33 & 10.00 & 5.00 & 10.00 & 0.00\\
           & \cellcolor{SeaGreen3!15}rules = 3 & \cellcolor{SeaGreen3!15}56.67 & \cellcolor{SeaGreen3!15}65.00 & \cellcolor{SeaGreen3!15}0.00 & \cellcolor{SeaGreen3!15}6.67 & \cellcolor{SeaGreen3!15}8.33 & \cellcolor{SeaGreen3!15}0.00 & \cellcolor{SeaGreen3!15}13.33 & \cellcolor{SeaGreen3!15}12.83 & \cellcolor{SeaGreen3!15}0.00\\
           \hdashline
           \multirow{3}{*}{Llama-3.1 405B} & rules = 1 & 70.00 & 45.83 & 70.00 & 30.00 & 9.33 & 10.00 & 10.00 & 1.67 & 10.00\\
           & rules = 2 & 50.00 & 33.33 & 10.00 & 25.00 & 11.39 & 0.00 & 10.00 & 3.00 & 0.00\\
           & \cellcolor{SeaGreen3!15}rules = 3 & \cellcolor{SeaGreen3!15}63.33 & \cellcolor{SeaGreen3!15}53.83 & \cellcolor{SeaGreen3!15}0.00 & \cellcolor{SeaGreen3!15}10.00 & \cellcolor{SeaGreen3!15}6.67 & \cellcolor{SeaGreen3!15}0.00 & \cellcolor{SeaGreen3!15}6.67 & \cellcolor{SeaGreen3!15}5.00 & \cellcolor{SeaGreen3!15}0.00\\
           \hdashline
           \multirow{3}{*}{GPT-4o} & rules = 1 & 30.00 & 12.50 & 30.00 & 30.00 & 10.83 & 10.00 & 10.00 & 5.00 & 10.00\\
           & rules = 2 & 75.00 & 63.17 & 60.00 & 20.00 & 7.42 & 0.00 & 15.00 & 6.35 & 0.00\\
           & \cellcolor{SeaGreen3!15}rules = 3 & \cellcolor{SeaGreen3!15}66.67 & \cellcolor{SeaGreen3!15}60.00 & \cellcolor{SeaGreen3!15}50.00 & \cellcolor{SeaGreen3!15}30.00 & \cellcolor{SeaGreen3!15}19.00 & \cellcolor{SeaGreen3!15}10.00 & \cellcolor{SeaGreen3!15}10.00 & \cellcolor{SeaGreen3!15}4.16 & \cellcolor{SeaGreen3!15}0.00\\
           \hdashline
           \multirow{3}{*}{DeepSeek-V3} & rules = 1 & 100.00 & 75.00 & 70.00 & 50.00 & 32.50 & 40.00 & 40.00 & 12.78 & 40.00\\
           & rules = 2 & 60.00 & 44.17 & 30.00 & 10.00 & 15.00 & 10.00 & 20.00 & 11.94 & 0.00\\
           & \cellcolor{SeaGreen3!15}rules = 3 & \cellcolor{SeaGreen3!15}83.33 & \cellcolor{SeaGreen3!15}77.67 & \cellcolor{SeaGreen3!15}50.00 & \cellcolor{SeaGreen3!15}20.00 & \cellcolor{SeaGreen3!15}12.92 & \cellcolor{SeaGreen3!15}0.00 & \cellcolor{SeaGreen3!15}23.33 & \cellcolor{SeaGreen3!15}13.97 & \cellcolor{SeaGreen3!15}0.00\\
           \hdashline
           \multirow{3}{*}{o1-mini} & rules = 1 & 90.00 & 90.00 & 90.00 & 70.00 & 55.00 & 40.00 & 10.00 & 10.00 & 10.00\\
           & rules = 2 & 80.00 & 80.00 & 80.00 & 60.00 & 60.00 & 50.00 & 65.00 & 53.83 & 10.00\\
           & \cellcolor{SeaGreen3!15}rules = 3 & \cellcolor{SeaGreen3!15}90.00 & \cellcolor{SeaGreen3!15}82.50 & \cellcolor{SeaGreen3!15}50.00 & \cellcolor{SeaGreen3!15}66.67 & \cellcolor{SeaGreen3!15}60.67 & \cellcolor{SeaGreen3!15}20.00 & \cellcolor{SeaGreen3!15}50.00 & \cellcolor{SeaGreen3!15}54.22 & \cellcolor{SeaGreen3!15}10.00\\
           \hdashline
           \multirow{3}{*}{o3-mini} & rules = 1 & 100.00 & 100.00 & 100.00 & 90/00 & 90.00 & 90.00 & 90.00 & 90.00 & 90.00\\
           & rules = 2 & 100.00 & 100.00 & 100.00 & 95.00 & 100.00 & 100.00 & 85.00 & 78.33 & 70.00\\
           & \cellcolor{SeaGreen3!15}rules = 3 & \cellcolor{SeaGreen3!15}\textbf{100.00} & \cellcolor{SeaGreen3!15}\textbf{100.00} & \cellcolor{SeaGreen3!15}\textbf{100.00} & \cellcolor{SeaGreen3!15}\textbf{86.67} & \cellcolor{SeaGreen3!15}\textbf{85.00} & \cellcolor{SeaGreen3!15}\textbf{80.00} & \cellcolor{SeaGreen3!15}\textbf{56.67} & \cellcolor{SeaGreen3!15}\textbf{50.33} & \cellcolor{SeaGreen3!15}\textbf{40.00}\\
           \midrule 
           \multicolumn{11}{c}{\textbf{vocab size = 3}}\\
         \midrule
           \multirow{3}{*}{Llama-3.3 70B} & rules = 1 & 50.00 & 50.00 & 50.00 & 20.00 & 12.50 & 10.00 & 20.00 & 13.33 & 10.00\\
           & rules = 2 & 35.00 & 33.67 & 10.00 & 20.00 & 6.93 & 10.00 & 25.00 & 15.00 & 0.00\\
           & \cellcolor{SeaGreen3!30}rules = 3 & \cellcolor{SeaGreen3!30}40.00 & \cellcolor{SeaGreen3!30}65.00 & \cellcolor{SeaGreen3!30}20.00 & \cellcolor{SeaGreen3!30}20.00 & \cellcolor{SeaGreen3!30}18.33 & \cellcolor{SeaGreen3!30}0.00 & \cellcolor{SeaGreen3!30}10.00 & \cellcolor{SeaGreen3!30}2.78 & \cellcolor{SeaGreen3!30}0.00\\
           \hdashline
           \multirow{3}{*}{Llama-3.1 405B} & rules = 1 & 60.00 & 45.00 & 40.00 & 10.00 & 3.33 & 10.00 & 10.00 & 1.11 & 0.00\\
           & rules = 2 & 30.00 & 20.00 & 0.00 & 15.00 & 13.33 & 0.00 & 5.00 & 0.53 & 0.00\\
           & \cellcolor{SeaGreen3!30}rules = 3 & \cellcolor{SeaGreen3!30}66.67 & \cellcolor{SeaGreen3!30}57.83 & \cellcolor{SeaGreen3!30}30.00 & \cellcolor{SeaGreen3!30}20.00 & \cellcolor{SeaGreen3!30}8.39 & \cellcolor{SeaGreen3!30}0.00 & \cellcolor{SeaGreen3!30}10.00 & \cellcolor{SeaGreen3!30}2.36 & \cellcolor{SeaGreen3!30}0.00\\
           \hdashline
           \multirow{3}{*}{GPT-4o} & rules = 1 & 40.00 & 27.50 & 40.00 & 20.00 & 8.33 & 20.00 & 40.00 & 11.00 & 30.00\\
           & rules = 2 & 55.00 & 46.50 & 10.00 & 45.00 & 25.67 & 0.00 & 30.00 & 6.50 & 10.00\\
           & \cellcolor{SeaGreen3!30}rules = 3 & \cellcolor{SeaGreen3!30}60.00 & \cellcolor{SeaGreen3!30}50.00 & \cellcolor{SeaGreen3!30}10.00 & \cellcolor{SeaGreen3!30}33.33 & \cellcolor{SeaGreen3!30}15.95 & \cellcolor{SeaGreen3!30}0.00 & \cellcolor{SeaGreen3!30}20.00 & \cellcolor{SeaGreen3!30}6.21 & \cellcolor{SeaGreen3!30}0.00\\
           \hdashline
           \multirow{3}{*}{DeepSeek-V3} & rules = 1 & 80.00 & 70.00 & 60.00 & 50.00 & 22.00 & 40.00 & 50.00 & 16.11 & 30.00\\
           & rules = 2 & 90.00 & 60.32 & 60.00 & 70.00 & 13.82 & 20.00 & 30.00 & 5.04 & 0.00\\
           & \cellcolor{SeaGreen3!30}rules = 3 & \cellcolor{SeaGreen3!30}66.67 & \cellcolor{SeaGreen3!30}53.50 & \cellcolor{SeaGreen3!30}40.00 & \cellcolor{SeaGreen3!30}23.33 & \cellcolor{SeaGreen3!30}16.42 & \cellcolor{SeaGreen3!30}0.00 & \cellcolor{SeaGreen3!30}30.00 & \cellcolor{SeaGreen3!30}7.54 & \cellcolor{SeaGreen3!30}\textbf{10.00}\\
           \hdashline
           \multirow{3}{*}{o1-mini} & rules = 1 & 100.00 & 95.00 & 90.00 & 80.00 & 63.33 & 70.00 & 30.00 & 17.50 & 30.00\\
           & rules = 2 & 90.00 & 83.33 & 80.00 & 70.00 & 49.42 & 40.00 & 35.00 & 29.52 & 20.00\\
           & \cellcolor{SeaGreen3!30}rules = 3 & \cellcolor{SeaGreen3!30}\textbf{96.67} & \cellcolor{SeaGreen3!30}\textbf{96.00} & \cellcolor{SeaGreen3!30}\textbf{90.00} & \cellcolor{SeaGreen3!30}70.00 & \cellcolor{SeaGreen3!30}56.15 & \cellcolor{SeaGreen3!30}30.00 & \cellcolor{SeaGreen3!30}50.00 & \cellcolor{SeaGreen3!30}33.58 & \cellcolor{SeaGreen3!30}0.00\\
           \hdashline
           \multirow{3}{*}{o3-mini} & rules = 1 & 100.00 & 100.00 & 100.00 & 90.00 & 90.00 & 90.00 & 80.00 & 80.00 & 80.00\\
           & rules = 2 & 100.00 & 100.00 & 100.00 & 90.00 & 75.15 & 70.00 & 80.00 & 72.50 & 50.00\\
           & \cellcolor{SeaGreen3!30}rules = 3 & \cellcolor{SeaGreen3!30}\textbf{96.67} & \cellcolor{SeaGreen3!30}94.17 & \cellcolor{SeaGreen3!30}\textbf{90.00} & \cellcolor{SeaGreen3!30}\textbf{96.67} & \cellcolor{SeaGreen3!30}\textbf{87.50} & \cellcolor{SeaGreen3!30}\textbf{90.00} & \cellcolor{SeaGreen3!30}\textbf{63.33} & \cellcolor{SeaGreen3!30}\textbf{68.43} & \cellcolor{SeaGreen3!30}\textbf{40.00}\\
           \midrule 
           \multicolumn{11}{c}{\textbf{vocab size = 4}}\\
         \midrule
           \multirow{3}{*}{Llama-3.3 70B} & rules = 1 & 50.00 & 29.00 & 30.00 & 20.00 & 13.33 & 10.00 & 10.00 & 10.00 & 10.00\\
           & rules = 2 & 50.00 & 50.00 & 10.00 & 20.00 & 15.96 & 0.00 & 0.00 & 0.00 & 0.00\\
           & \cellcolor{SeaGreen3!50}rules = 3 & \cellcolor{SeaGreen3!50}50.00 & \cellcolor{SeaGreen3!50}52.50 & \cellcolor{SeaGreen3!50}20.00 & \cellcolor{SeaGreen3!50}6.67 & \cellcolor{SeaGreen3!50}6.00 & \cellcolor{SeaGreen3!50}0.00 & \cellcolor{SeaGreen3!50}10.00 & \cellcolor{SeaGreen3!50}6.33 & \cellcolor{SeaGreen3!50}0.00\\
           \hdashline
           \multirow{3}{*}{Llama-3.1 405B} & rules = 1 & 60.00 & 34.50 & 30.00 & 10.00 & 5.00 & 10.00 & 10.00 & 5.00 & 10.00\\
           & rules = 2 & 50.00 & 29.00 & 0.00 & 10.00 & 3.13 & 0.00 & 10.00 & 2.90 & 0.00\\
           & \cellcolor{SeaGreen3!50}rules = 3 & \cellcolor{SeaGreen3!50}43.33 & \cellcolor{SeaGreen3!50}30.73 & \cellcolor{SeaGreen3!50}20.00 & \cellcolor{SeaGreen3!50}16.67 & \cellcolor{SeaGreen3!50}5.83 & \cellcolor{SeaGreen3!50}0.00 & \cellcolor{SeaGreen3!50}6.67 & \cellcolor{SeaGreen3!50}1.10 & \cellcolor{SeaGreen3!50}0.00\\
           \hdashline
           \multirow{3}{*}{GPT-4o} & rules = 1 & 40.00 & 35.00 & 30.00 & 60.00 & 25.33 & 40.00 & 40.00 & 12.50 & 10.00\\
           & rules = 2 & 75.00 & 45.50 & 30.00 & 55.00 & 20.47 & 10.00 & 20.00 & 9.44 & 0.00\\
           & \cellcolor{SeaGreen3!50}rules = 3 & \cellcolor{SeaGreen3!50}70.00 & \cellcolor{SeaGreen3!50}48.22 & \cellcolor{SeaGreen3!50}20.00 & \cellcolor{SeaGreen3!50}33.33 & \cellcolor{SeaGreen3!50}10.77 & \cellcolor{SeaGreen3!50}0.00 & \cellcolor{SeaGreen3!50}13.33 & \cellcolor{SeaGreen3!50}3.82 & \cellcolor{SeaGreen3!50}0.00\\
           \hdashline
           \multirow{3}{*}{DeepSeek-V3} & rules = 1 & 100.00 & 82.50 & 80.00 & 50.00 & 23.67 & 30.00 & 40.00 & 16.11 & 40.00\\
           & rules = 2 & 70.00 & 50.67 & 30.00 & 25.00 & 8.01 & 0.00 & 15.00 & 3.24 & 0.00\\
           & \cellcolor{SeaGreen3!50}rules = 3 & \cellcolor{SeaGreen3!50}60.00 & \cellcolor{SeaGreen3!50}48.36 & \cellcolor{SeaGreen3!50}30.00 & \cellcolor{SeaGreen3!50}50.00 & \cellcolor{SeaGreen3!50}12.81 & \cellcolor{SeaGreen3!50}20.00 & \cellcolor{SeaGreen3!50}23.33 & \cellcolor{SeaGreen3!50}2.73 & \cellcolor{SeaGreen3!50}0.00\\
           \hdashline
           \multirow{3}{*}{o1-mini} & rules = 1 & 90.00 & 85.00 & 90.00 & 50.00 & 38.33 & 30.00 & 40.00 & 28.33 & 20.00\\
           & rules = 2 & 100.00 & 93.33 & 80.00 & 60.00 & 36.92 & 20.00 & 50.00 & 31.92 & 10.00\\
           & \cellcolor{SeaGreen3!50}rules = 3 & \cellcolor{SeaGreen3!50}80.00 & \cellcolor{SeaGreen3!50}72.25 & \cellcolor{SeaGreen3!50}60.00 & \cellcolor{SeaGreen3!50}70.00 & \cellcolor{SeaGreen3!50}43.04 & \cellcolor{SeaGreen3!50}10.00 & \cellcolor{SeaGreen3!50}43.33 & \cellcolor{SeaGreen3!50}32.12 & \cellcolor{SeaGreen3!50}0.00\\
           \hdashline
           \multirow{3}{*}{o3-mini} & rules = 1 & 100.00 & 100.00 & 100.00 & 100.00 & 100.00 & 100.00 & 60.00 & 60.00 & 60.00\\
           & rules = 2 & 100.00 & 100.00 & 100.00 & 85.00 & 81.67 & 70.00 & 45.00 & 58.33 & 20.00\\
           & \cellcolor{SeaGreen3!50}rules = 3 & \cellcolor{SeaGreen3!50}\textbf{100.00} & \cellcolor{SeaGreen3!50}\textbf{100.00} & \cellcolor{SeaGreen3!50}\textbf{100.00} & \cellcolor{SeaGreen3!50}\textbf{76.67} & \cellcolor{SeaGreen3!50}\textbf{76.67} & \cellcolor{SeaGreen3!50}\textbf{70.00} & \cellcolor{SeaGreen3!50}\textbf{66.67} & \cellcolor{SeaGreen3!50}\textbf{69.17} & \cellcolor{SeaGreen3!50}\textbf{10.00}\\
           \bottomrule
    \end{tabular}
    }
    \caption{Left Output Strictly Local with sample size = 2}
    \label{tab:LOSL_main}
\end{table*}

\begin{table*}[t]
    \centering
    \renewcommand{\arraystretch}{1.1}
    \resizebox{14.5cm}{!}{
    \begin{tabular}{l c ccc ccc ccc}
        \toprule
          \multirow{2}{*}{\bf Models}& \multirow{2}{*}{\bf Settings}& \multicolumn{3}{c}{\bf k = 2} & \multicolumn{3}{c}{\bf k = 3} & \multicolumn{3}{c}{\bf k = 4}\\
          \cmidrule(lr){3-5} \cmidrule(lr){6-8}  \cmidrule(lr){9-11}
          & & recall  & precision & compatibility & recall  & precision & compatibility & recall  & precision & compatibility \\
         \midrule 
         \multicolumn{11}{c}{\textbf{vocab size = 2}} \\
         \midrule
           \multirow{3}{*}{Llama-3.3 70B} & rules = 1 & 40.00 & 40.00 & 40.00 & 20.00 & 10.00 & 10.00 & 20.00 & 10.00 & 10.00\\
           & rules = 2 & 40.00 & 40.00 & 30.00 & 15.00 & 18.33 & 10.00 & 20.00 & 18.67 & 10.00\\
           & \cellcolor{SeaGreen3!15}rules = 3 & \cellcolor{SeaGreen3!15}63.33 & \cellcolor{SeaGreen3!15}76.67 & \cellcolor{SeaGreen3!15}30.00 & \cellcolor{SeaGreen3!15}23.33 & \cellcolor{SeaGreen3!15}24.17 & \cellcolor{SeaGreen3!15}0.00 & \cellcolor{SeaGreen3!15}10.00 & \cellcolor{SeaGreen3!15}7.83 & \cellcolor{SeaGreen3!15}0.00\\
           \hdashline
           \multirow{3}{*}{Llama-3.1 405B} & rules = 1 & 20.00 & 8.33 & 0.00 & 40.00 & 18.33 & 0.00 & 0.00 & 0.00 & 0.00\\
           & rules = 2 & 60.00 & 51.67 & 40.00 & 25.00 & 12.50 & 0.00 & 10.00 & 6.25 & 10.00\\
           & \cellcolor{SeaGreen3!15}rules = 3 & \cellcolor{SeaGreen3!15}63.33 & \cellcolor{SeaGreen3!15}54.72 & \cellcolor{SeaGreen3!15}30.00 & \cellcolor{SeaGreen3!15}20.00 & \cellcolor{SeaGreen3!15}14.33 & \cellcolor{SeaGreen3!15}10.00 & \cellcolor{SeaGreen3!15}6.67 & \cellcolor{SeaGreen3!15}4.50 & \cellcolor{SeaGreen3!15}0.00\\
           \hdashline
           \multirow{3}{*}{GPT-4o} & rules = 1 & 50.00 & 30.00 & 30.00 & 10.00 & 5.00 & 10.00 & 0.00 & 0.00 & 0.00\\
           & rules = 2 & 70.00 & 61.67 & 60.00 & 45.00 & 16.93 & 10.00 & 30.00 & 15.83 & 20.00\\
           & \cellcolor{SeaGreen3!15}rules = 3 & \cellcolor{SeaGreen3!15}83.33 & \cellcolor{SeaGreen3!15}71.83 & \cellcolor{SeaGreen3!15}30.00 & \cellcolor{SeaGreen3!15}43.33 & \cellcolor{SeaGreen3!15}24.10 & \cellcolor{SeaGreen3!15}0.00 & \cellcolor{SeaGreen3!15}20.00 & \cellcolor{SeaGreen3!15}10.00 & \cellcolor{SeaGreen3!15}0.00\\
           \hdashline
           \multirow{3}{*}{DeepSeek-V3} & rules = 1 & 60.00 & 38.33 & 20.00 & 40.00 & 14.17 & 20.00 & 20.00 & 7.00 & 10.00\\
           & rules = 2 & 75.00 & 51.67 & 40.00 & 35.00 & 18.33 & 0.00 & 25.00 & 13.00 & 20.00\\
           & \cellcolor{SeaGreen3!15}rules = 3 & \cellcolor{SeaGreen3!15}86.67 & \cellcolor{SeaGreen3!15}72.56 & \cellcolor{SeaGreen3!15}50.00 & \cellcolor{SeaGreen3!15}40.00 & \cellcolor{SeaGreen3!15}27.00 & \cellcolor{SeaGreen3!15}0.00 & \cellcolor{SeaGreen3!15}23.33 & \cellcolor{SeaGreen3!15}3.68 & \cellcolor{SeaGreen3!15}0.00\\
           \hdashline
           \multirow{3}{*}{o1-mini} & rules = 1 & 50.00 & 38.33 & 40.00 & 20.00 & 5.83 & 0.00 & 20.00 & 11.67 & 10.00\\
           & rules = 2 & 55.00 & 38.33 & 30.00 & 35.00 & 17.50 & 0.00 & 15.00 & 8.70 & 0.00\\
           & \cellcolor{SeaGreen3!15}rules = 3 & \cellcolor{SeaGreen3!15}46.67 & \cellcolor{SeaGreen3!15}46.67 & \cellcolor{SeaGreen3!15}10.00 & \cellcolor{SeaGreen3!15}33.33 & \cellcolor{SeaGreen3!15}24.17 & \cellcolor{SeaGreen3!15}0.00 & \cellcolor{SeaGreen3!15}10.00 & \cellcolor{SeaGreen3!15}4.41 & \cellcolor{SeaGreen3!15}0.00\\
           \hdashline
           \multirow{3}{*}{o3-mini} & rules = 1 & 90.00 & 90.00 & 90.00 & 100.00 & 95.00 & 90.00 & 70.00 & 50.00 & 30.00\\
           & rules = 2 & 100.00 & 100.00 & 100.00 & 90.00 & 85.00 & 60.00 & 45.00 & 41.67 & 20.00\\
           & \cellcolor{SeaGreen3!15}rules = 3 & \cellcolor{SeaGreen3!15}\textbf{96.67} & \cellcolor{SeaGreen3!15}\textbf{92.67} & \cellcolor{SeaGreen3!15}\textbf{80.00} & \cellcolor{SeaGreen3!15}\textbf{86.67} & \cellcolor{SeaGreen3!15}\textbf{78.33} & \cellcolor{SeaGreen3!15}\textbf{50.00} & \cellcolor{SeaGreen3!15}\textbf{46.67} & \cellcolor{SeaGreen3!15}\textbf{46.67} & \cellcolor{SeaGreen3!15}\textbf{30.00}\\
         \midrule
           \multirow{3}{*}{Llama-3.3 70B} & rules = 1 & 40.00 & 40.00 & 40.00 & 30.00 & 25.00 & 30.00 & 30.00 & 7.26 & 0.00\\
           & rules = 2 & 30.00 & 60.00 & 10.00 & 20.00 & 10.93 & 0.00 & 25.00 & 19.17 & 10.00\\
           & \cellcolor{SeaGreen3!30}rules = 3 & \cellcolor{SeaGreen3!30}30.00 & \cellcolor{SeaGreen3!30}45.00 & \cellcolor{SeaGreen3!30}0.00 & \cellcolor{SeaGreen3!30}26.67 & \cellcolor{SeaGreen3!30}22.67 & \cellcolor{SeaGreen3!30}10.00 & \cellcolor{SeaGreen3!30}10.00 & \cellcolor{SeaGreen3!30}4.58 & \cellcolor{SeaGreen3!30}0.00\\
           \hdashline
           \multirow{3}{*}{Llama-3.1 405B} & rules = 1 & 20.00 & 11.43 & 10.00 & 10.00 & 5.00 & 0.00 & 20.00 & 4.17 & 0.00\\
           & rules = 2 & 30.00 & 20.83 & 10.00 & 15.00 & 4.17 & 0.00 & 10.00 & 3.41 & 0.00\\
           & \cellcolor{SeaGreen3!30}rules = 3 & \cellcolor{SeaGreen3!30}16.67 & \cellcolor{SeaGreen3!30}9.11 & \cellcolor{SeaGreen3!30}0.00 & \cellcolor{SeaGreen3!30}10.00 & \cellcolor{SeaGreen3!30}4.75 & \cellcolor{SeaGreen3!30}0.00 & \cellcolor{SeaGreen3!30}0.00 & \cellcolor{SeaGreen3!30}0.00 & \cellcolor{SeaGreen3!30}0.00\\
            \hdashline
           \multirow{3}{*}{GPT-4o} & rules = 1 & 60.00 & 50.00 & 40.00 & 50.00 & 27.50 & 20.00 & 50.00 & 15.33 & 20.00\\
           & rules = 2 & 60.00 & 41.67 & 30.00 & 50.00 & 27.50 & 20.00 & 30.00 & 8.82 & 0.00\\
           & \cellcolor{SeaGreen3!30}rules = 3 & \cellcolor{SeaGreen3!30}66.67 & \cellcolor{SeaGreen3!30}57.83 & \cellcolor{SeaGreen3!30}30.00 & \cellcolor{SeaGreen3!30}40.00 & \cellcolor{SeaGreen3!30}21.88 & \cellcolor{SeaGreen3!30}0.00 & \cellcolor{SeaGreen3!30}13.33 & \cellcolor{SeaGreen3!30}6.33 & \cellcolor{SeaGreen3!30}0.00\\
           \hdashline
           \multirow{3}{*}{DeepSeek-V3} & rules = 1 & 80.00 & 60.83 & 60.00 & 50.00 & 29.17 & 40.00 & 40.00 & 9.42 & 20.00\\
           & rules = 2 & 80.00 & 63.19 & 50.00 & 55.00 & 26.67 & 20.00 & 40.00 & 10.10 & 0.00\\
           & \cellcolor{SeaGreen3!30}rules = 3 & \cellcolor{SeaGreen3!30}70.00 & \cellcolor{SeaGreen3!30}42.95 & \cellcolor{SeaGreen3!30}50.00 & \cellcolor{SeaGreen3!30}23.33 & \cellcolor{SeaGreen3!30}15.28 & \cellcolor{SeaGreen3!30}0.00 & \cellcolor{SeaGreen3!30}20.00 & \cellcolor{SeaGreen3!30}3.56 & \cellcolor{SeaGreen3!30}0.00\\
           \hdashline
           \multirow{3}{*}{o1-mini} & rules = 1 & 60.00 & 60.00 & 60.00 & 20.00 & 13.33 & 10.00 & 20.00 & 13.33 & 10.00\\
           & rules = 2 & 15.00 & 15.00 & 10.00 & 40.00 & 25.00 & 0.00 & 30.00 & 19.50 & 0.00\\
           & \cellcolor{SeaGreen3!30}rules = 3 & \cellcolor{SeaGreen3!30}53.33 & \cellcolor{SeaGreen3!30}45.83 & \cellcolor{SeaGreen3!30}20.00 & \cellcolor{SeaGreen3!30}30.00 & \cellcolor{SeaGreen3!30}17.63 & \cellcolor{SeaGreen3!30}0.00 & \cellcolor{SeaGreen3!30}13.33 & \cellcolor{SeaGreen3!30}8.43 & \cellcolor{SeaGreen3!30}0.00\\
           \hdashline
           \multirow{3}{*}{o3-mini} & rules = 1 & 90.00 & 90.00 & 90.00 & 100.00 & 95.00 & 100.00 & 50.00 & 40.00 & 30.00\\
           & rules = 2 & 90.00 & 83.33 & 70.00 & 80.00 & 61.67 & 30.00 & 60.00 & 59.50 & 40.00\\
           & \cellcolor{SeaGreen3!30}rules = 3 & \cellcolor{SeaGreen3!30}\textbf{100.00} & \cellcolor{SeaGreen3!30}\textbf{100.00} & \cellcolor{SeaGreen3!30}\textbf{100.00} & \cellcolor{SeaGreen3!30}\textbf{83.33} & \cellcolor{SeaGreen3!30}\textbf{64.11} & \cellcolor{SeaGreen3!30}\textbf{50.00} & \cellcolor{SeaGreen3!30}\textbf{43.33} & \cellcolor{SeaGreen3!30}\textbf{41.83} & \cellcolor{SeaGreen3!30}\textbf{20.00}\\
           \midrule 
           \multicolumn{11}{c}{\textbf{vocab size = 4}}\\
         \midrule
           \multirow{3}{*}{Llama-3.3 70B} & rules = 1 & 40.00 & 25.00 & 30.00 & 40.00 & 23.33 & 20.00 & 10.00 & 10.00 & 10.00\\
           & rules = 2 & 45.00 & 47.33 & 10.00 & 50.00 & 45.83 & 10.00 & 5.00 & 1.67 & 0.00\\
           & \cellcolor{SeaGreen3!50}rules = 3 & \cellcolor{SeaGreen3!50}33.33 & \cellcolor{SeaGreen3!50}39.50 & \cellcolor{SeaGreen3!50}10.00 & \cellcolor{SeaGreen3!50}30.00 & \cellcolor{SeaGreen3!50}23.85 & \cellcolor{SeaGreen3!50}0.00 & \cellcolor{SeaGreen3!50}10.00 & \cellcolor{SeaGreen3!50}10.83 & \cellcolor{SeaGreen3!50}0.00\\
           \hdashline
           \multirow{3}{*}{Llama-3.1 405B} & rules = 1 & 10.00 & 10.00 & 10.00 & 0.00 & 0.00 & 0.00 & 0.00 & 0.00 & 0.00\\
           & rules = 2 & 15.00 & 12.00 & 0.00 & 10.00 & 13.33 & 0.00 & 5.00 & 0.26 & 0.00\\
           & \cellcolor{SeaGreen3!50}rules = 3 & \cellcolor{SeaGreen3!50}33.33 & \cellcolor{SeaGreen3!50}22.67 & \cellcolor{SeaGreen3!50}0.00 & \cellcolor{SeaGreen3!50}3.33 & \cellcolor{SeaGreen3!50}16.67 & \cellcolor{SeaGreen3!50}0.00 & \cellcolor{SeaGreen3!50}13.33 & \cellcolor{SeaGreen3!50}1.85 & \cellcolor{SeaGreen3!50}0.00\\
           \hdashline
           \multirow{3}{*}{GPT-4o} & rules = 1 & 70.00 & 49.17 & 50.00 & 60.00 & 19.50 & 50.00 & 50.00 & 23.10 & 10.00\\
           & rules = 2 & 80.00 & 78.10 & 40.00 & 40.00 & 16.02 & 0.00 & 20.00 & 10.00 & 0.00\\
           & \cellcolor{SeaGreen3!50}rules = 3 & \cellcolor{SeaGreen3!50}66.67 & \cellcolor{SeaGreen3!50}43.00 & \cellcolor{SeaGreen3!50}0.00 & \cellcolor{SeaGreen3!50}20.00 & \cellcolor{SeaGreen3!50}11.02 & \cellcolor{SeaGreen3!50}0.00 & \cellcolor{SeaGreen3!50}16.67 & \cellcolor{SeaGreen3!50}6.73 & \cellcolor{SeaGreen3!50}0.00\\
           \hdashline
           \multirow{3}{*}{DeepSeek-V3} & rules = 1 & 80.00 & 65.00 & 70.00 & 50.00 & 18.83 & 20.00 & 60.00 & 19.77 & 40.00\\
           & rules = 2 & 70.00 & 59.17 & 30.00 & 45.00 & 27.79 & 20.00 & 10.00 & 0.88 & 0.00\\
           & \cellcolor{SeaGreen3!50}rules = 3 & \cellcolor{SeaGreen3!50}73.33 & \cellcolor{SeaGreen3!50}60.17 & \cellcolor{SeaGreen3!50}40.00 & \cellcolor{SeaGreen3!50}43.33 & \cellcolor{SeaGreen3!50}13.45 & \cellcolor{SeaGreen3!50}0.00 & \cellcolor{SeaGreen3!50}3.33 & \cellcolor{SeaGreen3!50}0.25 & \cellcolor{SeaGreen3!50}0.00\\
           \hdashline
           \multirow{3}{*}{o1-mini} & rules = 1 & 70.00 & 53.33 & 40.00 & 30.00 & 18.33 & 10.00 & 40.00 & 40.00 & 40.00\\
           & rules = 2 & 70.00 & 66.67 & 50.00 & 50.00 & 47.50 & 20.00 & 40.00 & 34.00 & 0.00\\
           & \cellcolor{SeaGreen3!50}rules = 3 & \cellcolor{SeaGreen3!50}90.00 & \cellcolor{SeaGreen3!50}79.17 & \cellcolor{SeaGreen3!50}50.00 & \cellcolor{SeaGreen3!50}23.33 & \cellcolor{SeaGreen3!50}23.33 & \cellcolor{SeaGreen3!50}0.00 & \cellcolor{SeaGreen3!50}26.67 & \cellcolor{SeaGreen3!50}17.58 & \cellcolor{SeaGreen3!50}0.00\\
           \hdashline
           \multirow{3}{*}{o3-mini} & rules = 1 & 90.00 & 90.00 & 90.00 & 100.00 & 93.33 & 90.00 & 50.00 & 50.00 & 50.00\\
           & rules = 2 & 100.00 & 100.00 & 100.00 & 80.00 & 75.00 & 60.00 & 70.00 & 69.17 & 40.00\\
           & \cellcolor{SeaGreen3!50}rules = 3 & \cellcolor{SeaGreen3!50}\textbf{100.00} & \cellcolor{SeaGreen3!50}\textbf{100.00} & \cellcolor{SeaGreen3!50}\textbf{100.00} & \cellcolor{SeaGreen3!50}\textbf{76.67} & \cellcolor{SeaGreen3!50}\textbf{78.33} & \cellcolor{SeaGreen3!50}\textbf{50.00} & \cellcolor{SeaGreen3!50}\textbf{63.33} & \cellcolor{SeaGreen3!50}\textbf{62.00} & \cellcolor{SeaGreen3!50}\textbf{30.00}\\
           \bottomrule
    \end{tabular}
    }
    \caption{Right Output Strictly Local with sample size = 2}
    \label{tab:ROSL_main}
\end{table*}

\begin{table*}[t]
    \centering
    \renewcommand{\arraystretch}{1.1}
    \resizebox{14.5cm}{!}{
    \begin{tabular}{l c ccc ccc ccc}
        \toprule
          \multirow{2}{*}{\bf Models}& \multirow{2}{*}{\bf Settings}& \multicolumn{3}{c}{\bf k = 2} & \multicolumn{3}{c}{\bf k = 3} & \multicolumn{3}{c}{\bf k = 4}\\
          \cmidrule(lr){3-5} \cmidrule(lr){6-8}  \cmidrule(lr){9-11}
          & & recall  & precision & compatibility & recall  & precision & compatibility & recall  & precision & compatibility \\
         \midrule \multicolumn{11}{c}{\textbf{vocab size = 2}} \\
         \midrule
           \multirow{4}{*}{rules = 1} & 0-shot & 60.00 & 55.00 & 60.00 & 30.00 & 23.33 & 20.00 & 10.00 & 10.00 & 10.00  \\
           & 1-shot & 60.00 & 60.00 & 60.00 & 50.00 & 35.00 & 40.00 & 10.00 & 5.00 & 0.00\\
           & 2-shot & 70.00 & 70.00 & 70.00 & 70.00 & 50.00 & 60.00 & 20.00 & 10.00 & 10.00\\
           & 3-shot & 80.00 & 80.00 & 80.00 & 60.00 & 55.00 & 60.00 & 10.00 & 5.00 & 10.00\\
           \hdashline
           \multirow{4}{*}{rules = 2} &  0-shot & 60.00 & 65.00 & 50.00 & 45.00 & 60.00 & 30.00 & 15.00 & 8.25 & 0.00\\
           & 1-shot & 65.00 & 70.00 & 60.00 & 40.00 & 53.00 & 30.00 & 20.00 & 18.33 & 10.00\\
           & 2-shot & 85.00 & 85.00 & 80.00 & 45.00 & 56.67 & 20.00 & 25.00 & 23.33 & 0.00\\
           & 3-shot & 60.00 & 65.00 & 40.00 & 35.00 & 36.67 & 10.00 & 20.00 & 20.00 & 0.00\\
           \hdashline
          \multirow{4}{*}{rules = 3} & 0-shot & 53.33 & 68.33 & 20.00 & 30.00 & 46.67 & 10.00 & 16.67 & 8.54 & 0.00\\
          & 1-shot & 76.67 & 83.33 & 60.00 & 43.33 & 57.67 & 0.00 & 20.00 & 18.33 & 0.00\\
           & 2-shot & 86.67 & 86.67 & 60.00 & 26.67 & 28.33 & 0.00 & 13.33 & 16.67 & 0.00\\
           & 3-shot & 90.00 & 93.33 & 70.00 & 46.67 & 52.50 & 20.00 & 16.67 & 21.17 & 0.00\\
           \midrule \multicolumn{11}{c}{\textbf{vocab size = 3}} \\
           \midrule
           \multirow{4}{*}{rules = 1} & 0-shot & 70.00 & 60.00 & 60.00 & 20.00 & 20.00 & 20.00 & 20.00 & 8.33 & 10.00\\
           & 1-shot & 90.00 & 90.00 & 90.00 & 50.00 & 50.00 & 50.00 & 30.00 & 30.00 & 30.00\\
           & 2-shot & 70.00  & 70.00  & 70.00 & 30.00 & 20.00 & 20.00 & 10.00 & 10.00 & 10.00\\
           & 3-shot & 40.00 & 40.00 & 40.00 & 40.00 & 35.00 & 40.00 & 30.00 & 18.33 & 20.00\\
           \hdashline
           \multirow{4}{*}{rules = 2} &  0-shot & 85.00 & 83.33 & 60.00 & 10.00 & 7.50 & 0.00 & 5.00 & 2.50 & 0.00\\
           & 1-shot & 90.00 & 95.00 & 80.00 & 10.00 & 13.33 & 0.00 & 5.00 & 2.50 & 0.00\\
           & 2-shot & 65.00 & 65.00 & 40.00 & 30.00 & 28.33 & 10.00 & 5.00 & 2.00 & 0.00\\
           & 3-shot & 65.00 & 75.00 & 30.00 & 5.00 & 3.33 & 0.00 & 5.00 & 2.50 & 0.00\\
           \hdashline
          \multirow{4}{*}{rules = 3} & 0-shot & 66.67 & 74.17 & 20.00 & 33.33 & 35.36 & 0.00 & 6.67 & 3.43 & 0.00\\
          & 1-shot & 70.00 & 69.17 & 40.00 & 20.00 & 22.50 & 0.00 & 10.00 & 13.33 & 0.00\\
           & 2-shot & 76.67 & 76.67 & 50.00 & 33.33 & 43.33 & 0.00 & 10.00 & 13.33 & 0.00\\
           & 3-shot & 60.00 & 60.83 & 10.00 & 23.33 & 31.67 & 0.00 & 16.67 & 19.76 & 0.00\\
           \midrule \multicolumn{11}{c}{\textbf{vocab size = 4}} \\
           \midrule
           \multirow{4}{*}{rules = 1} & 0-shot & 60.00 & 60.00 & 60.00 & 30.00 & 30.00 & 30.00 & 10.00 & 10.00 & 10.00\\
           & 1-shot & 40.00 & 40.00 & 40.00 & 50.00 & 31.67 & 50.00 & 20.00 & 13.33 & 20.00\\
           & 2-shot & 70.00 & 57.00 & 60.00 & 30.00 & 25.00 & 30.00 & 10.00 & 5.00 & 0.00\\
           & 3-shot & 60.00 & 60.00 & 60.00 & 20.00 & 15.00 & 20.00 & 10.00 & 10.00 & 10.00\\
           \hdashline
           \multirow{4}{*}{rules = 2} &  0-shot & 40.00 & 40.00 & 30.00 & 15.00 & 11.67 & 0.00 & 0.00 & 0.00 & 0.00\\
           & 1-shot & 60.00 & 60.00 & 30.00 & 15.00 & 23.33 & 0.00 & 0.00 & 0.00 & 0.00\\
           & 2-shot & 80.00 & 90.00 & 60.00 & 15.00 & 12.50 & 0.00 & 15.00 & 10.67 & 0.00\\
           & 3-shot & 70.00 & 70.00 & 70.00 & 15.00 & 18.33 & 0.00 & 10.00 & 10.00 & 0.00\\
           \hdashline
          \multirow{4}{*}{rules = 3} & 0-shot & 53.33 & 68.33 & 20.00 & 6.67 & 5.00 & 0.00 & 10.00 & 5.32 & 0.00\\
          & 1-shot & 66.67 & 70.83 & 30.00 & 30.00 & 47.50 & 0.00 & 3.00 & 5.00 & 0.00\\
           & 2-shot & 66.67 & 73.33 & 30.00 & 30.00 & 55.00 & 0.00 & 3.33 & 3.33 & 0.00\\
           & 3-shot & 60.00 & 71.67 & 10.00 & 20.00 & 39.24 & 0.00 & 0.00 & 0.00 & 0.00\\
           \bottomrule
    \end{tabular}
    }
    \caption{Input Strictly Local with sample size = 2 with few-shot example}
    \label{tab:few_shot_ISL}
\end{table*}

\begin{table*}[t]
    \centering
    \renewcommand{\arraystretch}{1.1}
    \resizebox{14.5cm}{!}{
    \begin{tabular}{l c ccc ccc ccc}
        \toprule
          \multirow{2}{*}{\bf Models}& \multirow{2}{*}{\bf Settings}& \multicolumn{3}{c}{\bf k = 2} & \multicolumn{3}{c}{\bf k = 3} & \multicolumn{3}{c}{\bf k = 4}\\
          \cmidrule(lr){3-5} \cmidrule(lr){6-8}  \cmidrule(lr){9-11}
          & & recall  & precision & compatibility & recall  & precision & compatibility & recall  & precision & compatibility \\
         \midrule \multicolumn{11}{c}{\textbf{vocab size = 2}} \\
         \midrule
           \multirow{4}{*}{rules = 1} & 0-shot & 50.00 & 45.00 & 50.00 & 0.00 & 0.00 & 0.00 & 0.00 & 0.00 & 0.00\\
           & 1-shot & 80.00 & 80.00 & 80.00 & 40.00 & 33.33 & 30.00 & 20.00 & 5.00 & 20.00\\
           & 2-shot & 80.00 & 75.00 & 70.00 & 30.00 & 25.00 & 30.00 & 30.00 & 13.33 & 10.00\\
           & 3-shot & 80.00 & 80.00 & 80.00 & 20.00 & 15.00 & 20.00 & 20.00 & 15.00 & 20.00\\
           \hdashline
           \multirow{4}{*}{rules = 2} &  0-shot & 25.00& 25.00 & 25.00 & 10.00 & 8.33 & 10.00 & 5.00 & 10.00 & 0.00\\
           & 1-shot & 80.00 & 85.00 & 70.00 & 30.00 & 30.83 & 10.00 & 30.00 & 19.00 & 10.00\\
           & 2-shot & 85.00 & 85.00 & 80.00 & 20.00 & 21.67 & 10.00 & 25.00 & 27.90 & 0.00\\
           & 3-shot & 75.00 & 80.00 & 60.00 & 25.00 & 20.83 & 10.00 & 20.00 & 20.83 & 0.00\\
           \hdashline
          \multirow{4}{*}{rules = 3} & 0-shot & 56.67 & 65.00 & 0.00 & 6.67 & 8.33 & 0.00 & 13.33 & 12.83 & 0.00\\
          & 1-shot & 80.00 & 80.00 & 80.00 & 40.00 & 42.00 & 0.00 & 10.00 & 11.67 & 0.00\\
           & 2-shot & 80.00 & 75.00 & 70.00 & 33.33 & 48.33 & 0.00 & 13.33 & 28.33 & 0.00\\
           & 3-shot & 80.00 & 80.00 & 80.00 & 33.33 & 39.17 & 0.00 & 16.67 & 26.67 & 0.00\\
           \midrule \multicolumn{11}{c}{\textbf{vocab size = 3}} \\
           \midrule
           \multirow{4}{*}{rules = 1} & 0-shot & 50.00 & 50.00 & 50.00 & 20.00 & 12.50 & 10.00 & 20.00 & 13.33 & 10.00\\
           & 1-shot & 100.00 & 100.00 & 100.00 & 40.00 & 23.33 & 40.00 & 0.00 & 0.00 & 0.00\\
           & 2-shot & 70.00 & 70.00 & 70.00 & 30.00 & 23.33 & 20.00 & 10.00 & 5.00 & 0.00\\
           & 3-shot & 80.00 & 75.00 & 80.00 & 20.00 & 20.00 & 20.00 & 20.00 & 20.00 & 20.00\\
           \hdashline
           \multirow{4}{*}{rules = 2} &  0-shot & 35.00 & 33.67 & 10.00 & 20.00 & 6.93 & 10.00 & 25.00 & 15.00 & 0.00\\
           & 1-shot & 80.00 & 76.67 & 80.00 & 30.00 & 30.33 & 10.00 & 10.00 & 15.00 & 0.00\\
           & 2-shot & 50.00 & 55.00 & 30.00 & 30.00 & 38.33 & 0.00 & 25.00 & 21.67 & 0.00\\
           & 3-shot & 70.00 & 70.00 & 70.00 & 20.00 & 30.00 & 0.00 & 20.00 & 35.00 & 0.00\\
           \hdashline
          \multirow{4}{*}{rules = 3} & 0-shot & 40.00 & 65.00 & 20.00 & 20.00 & 18.33 & 0.00 & 10.00 & 2.78 & 0.00\\
          & 1-shot & 70.00 & 66.67 & 40.00 & 30.00 & 24.83 & 0.00 & 10.00 & 20.30 & 10.00\\
           & 2-shot & 83.33 & 90.00 & 60.00 & 33.33 & 40.83 & 0.00 & 30.00 & 43.33 & 0.00\\
           & 3-shot & 70.00 & 78.33 & 30.00 & 23.33 & 44.50 & 0.00 & 16.67 & 18.33 & 0.00\\
           \midrule \multicolumn{11}{c}{\textbf{vocab size = 4}} \\
           \midrule
           \multirow{4}{*}{rules = 1} & 0-shot & 50.00 & 29.00 & 30.00 & 20.00 & 13.33 & 10.00 & 10.00 & 10.00 & 10.00\\
           & 1-shot & 50.00 & 50.00 & 50.00 & 50.00 & 50.00 & 50.00 & 20.00 & 15.00 & 20.00\\
           & 2-shot & 60.00 & 60.00 & 60.00 & 10.00 & 10.00 & 10.00 & 20.00 & 20.00 & 20.00\\
           & 3-shot & 20.00 & 20.00 & 20.00 & 30.00 & 25.00 & 30.00 & 20.00 & 20.00 & 20.00\\
           \hdashline
           \multirow{4}{*}{rules = 2} &  0-shot & 50.00 & 50.00 & 10.00 & 20.00 & 15.96 & 0.00 & 0.00 & 0.00 & 0.00\\
           & 1-shot & 55.00 & 56.67 & 30.00 & 20.00 & 28.33 & 0.00 & 0.00 & 0.00 & 0.00\\
           & 2-shot & 55.00 & 50.00 & 20.00 & 35.00 & 55.00 & 10.00 & 5.00 & 10.00 & 0.00\\
           & 3-shot & 10.00 & 20.00 & 0.00 & 30.00 & 33.33 & 10.00 & 10.00 & 20.00 & 0.00\\
           \hdashline
          \multirow{4}{*}{rules = 3} & 0-shot & 50.00 & 52.50 & 20.00 & 6.67 & 6.00 & 0.00 & 10.00 & 6.33 & 0.00\\
          & 1-shot & 56.67 & 68.33 & 20.00 & 20.00 & 36.25 & 0.00 & 16.7 & 22.83 & 0.00\\
           & 2-shot & 66.67 & 67.50 & 50.00 & 16.67 & 26.67 & 0.00 & 6.67 & 15.00 & 0.00\\
           & 3-shot & 3.33 & 3.33 & 0.00 & 23.33 & 40.83 & 0.00 & 3.33 & 3.33 & 0.00\\
           \bottomrule
    \end{tabular}
    }
    \caption{Left Output Strictly Local with sample size = 2 with few-shot example}
    \label{tab:few_shot_LOSL}
\end{table*}

\begin{table*}[t]
    \centering
    \renewcommand{\arraystretch}{1.1}
    \resizebox{14.5cm}{!}{
    \begin{tabular}{l c ccc ccc ccc}
        \toprule
          \multirow{2}{*}{\bf Models}& \multirow{2}{*}{\bf Settings}& \multicolumn{3}{c}{\bf k = 2} & \multicolumn{3}{c}{\bf k = 3} & \multicolumn{3}{c}{\bf k = 4}\\
          \cmidrule(lr){3-5} \cmidrule(lr){6-8}  \cmidrule(lr){9-11}
          & & recall  & precision & compatibility & recall  & precision & compatibility & recall  & precision & compatibility \\
         \midrule \multicolumn{11}{c}{\textbf{vocab size = 2}} \\
         \midrule
           \multirow{4}{*}{rules = 1} & 0-shot & 40.00 & 40.00 & 40.00 & 20.00 & 10.00 & 10.00 & 20.00 & 10.00 & 10.00\\
           & 1-shot & 60.00 & 60.00 & 60.00 & 50.00 & 40.00 & 50.00 & 30.00 & 18.33 & 10.00\\
           & 2-shot & 60.00 & 60.00 & 60.00 & 40.00 & 35.00 & 40.00 & 30.00 & 25.00 & 20.00\\
           & 3-shot & 60.00 & 60.00 & 60.00 & 50.00 & 40.00 & 50.00 & 10.00 & 2.50 & 10.00\\
           \hdashline
           \multirow{4}{*}{rules = 2} &  0-shot & 40.00 & 40.00 & 30.00 & 15.00 & 18.33 & 10.00 & 20.00 & 18.67 & 10.00\\
           & 1-shot & 60.00 & 60.00 & 50.00 & 40.00 & 42.50 & 20.00 & 30.00 & 33.33 & 0.00\\
           & 2-shot & 70.00 & 80.00 & 60.00 & 45.00 & 43.33 & 30.00 & 20.00 & 27.50 & 0.00\\
           & 3-shot & 90.00 & 90.00 & 90.00 & 45.00 & 41.67 & 30.00 & 15.00 & 15.00 & 0.00\\
           \hdashline
          \multirow{4}{*}{rules = 3} & 0-shot & 63.33 & 76.67 & 30.00 & 23.33 & 24.17 & 0.00 & 10.00 & 7.83 & 0.00\\
          & 1-shot & 60.00 & 60.00 & 60.00 & 40.00 & 47.83 & 0.00 & 20.00 & 24.17 & 0.00\\
           & 2-shot & 60.00 & 60.00 & 60.00 & 50.00 & 48.33 & 20.00 & 16.67 & 18.33 & 0.00\\
           & 3-shot & 60.00 & 60.00 & 60.00 & 43.33 & 46.67 & 10.00 & 16.67 & 25.00 & 0.00\\
           \midrule \multicolumn{11}{c}{\textbf{vocab size = 3}} \\
           \midrule
           \multirow{4}{*}{rules = 1} & 0-shot & 40.00 & 40.00 & 40.00 & 30.00 & 25.00 & 30.00 & 30.00 & 7.26 & 0.00\\
           & 1-shot & 70.00 & 65.00 & 60.00 & 40.00 & 35.00 & 40.00 & 20.00 & 15.00 & 10.00\\
           & 2-shot & 80.00 & 75.00 & 70.00 & 50.00 & 50.00 & 50.00 & 30.00 & 18.33 & 30.00\\
           & 3-shot & 70.00 & 65.00 & 60.00 & 40.00 & 35.00 & 40.00 & 20.00 & 15.00 & 10.00\\
           \hdashline
           \multirow{4}{*}{rules = 2} &  0-shot & 30.00 & 60.00 & 10.00 & 20.00 & 10.93 & 0.00 & 25.00 & 19.17 & 10.00\\
           & 1-shot & 65.00 & 70.00 & 40.00 & 30.00 & 45.00 & 10.00 & 15.00 & 11.67 & 0.00\\
           & 2-shot & 60.00 & 70.00 & 40.00 & 50.00 & 44.17 & 10.00 & 20.00 & 20.00 & 0.00\\
           & 3-shot & 65.00 & 70.00 & 40.00 & 30.00 & 45.00 & 10.00 & 15.00 & 11.67 & 0.00\\
           \hdashline
          \multirow{4}{*}{rules = 3} & 0-shot & 30.00 & 45.00 & 0.00 & 26.67 & 22.67 & 10.00 & 10.00 & 4.58 & 0.00\\
          & 1-shot & 80.00 & 80.83 & 40.00 & 30.00 & 37.59 & 0.00 & 13.33 & 11.00 & 0.00\\
           & 2-shot & 73.33 & 71.67 & 30.00 & 26.67 & 34.50 & 0.00 & 33.33 & 47.00 & 0.00\\
           & 3-shot & 80.00 & 80.83 & 40.00 & 30.00 & 37.60 & 0.00 & 13.33 & 11.00 & 0.00\\
           \midrule \multicolumn{11}{c}{\textbf{vocab size = 4}} \\
           \midrule
           \multirow{4}{*}{rules = 1} & 0-shot & 40.00 & 25.00 & 30.00 & 40.00 & 23.33 & 20.00 & 10.00 & 10.00 & 10.00\\
           & 1-shot & 60.00 & 60.00 & 60.00 & 50.00 & 31.67 & 50.00 & 10.00 & 10.00 & 10.00\\
           & 2-shot & 80.00 & 68.33 & 80.00 & 60.00 & 33.33 & 30.00 & 20.00 & 20.00 & 20.00\\
           & 3-shot & 70.00 & 70.00 & 70.00 & 40.00 & 19.50 & 40.00 & 20.00 & 20.00 & 20.00\\
           \hdashline
           \multirow{4}{*}{rules = 2} &  0-shot & 45.00 & 47.33 & 10.00 & 50.00 & 45.83 & 10.00 & 5.00 & 1.67 & 0.00\\
           & 1-shot & 80.00 & 70.00 & 40.00 & 45.00 & 61.67 & 10.00 & 5.00 & 10.00 & 0.00\\
           & 2-shot & 65.00 & 65.00 & 40.00 & 45.00 & 52.50 & 20.00 & 0.00 & 0.00 & 0.00\\
           & 3-shot & 75.00 & 75.00 & 70.00 & 45.00 & 48.33 & 20.00 & 15.00 & 30.00 & 0.00\\
           \hdashline
          \multirow{4}{*}{rules = 3} & 0-shot & 33.33 & 39.50 & 10.00 & 30.00 & 23.85 & 0.00 & 10.00 & 10.83 & 0.00\\
          & 1-shot & 66.67 & 71.83 & 30.00 & 23.33 & 30.83 & 0.00 & 3.33 & 2.50 & 0.00\\
           & 2-shot & 83.33 & 86.50 & 50.00 & 26.67 & 37.50 & 0.00 & 13.33 & 28.83 & 0.00\\
           & 3-shot & 76.67 & 80.00 & 50.00 & 40.00 & 51.67 & 10.00 & 13.33 & 20.00 & 0.00\\
           \bottomrule
    \end{tabular}
    }
    \caption{Right Output Strictly Local with sample size = 2 with few-shot example}
    \label{tab:few_shot_ROSL}
\end{table*}

\end{document}